%% file: main.tex
\documentclass[10pt,twocolumn,letterpaper]{article}

\usepackage[accsupp]{axessibility}  

\usepackage{iccv}              
\input{math_commands.tex}

\definecolor{iccvblue}{rgb}{0.21,0.49,0.74}
\usepackage[pagebackref,breaklinks,colorlinks,allcolors=iccvblue]{hyperref}
\usepackage{multirow}
\usepackage{float}
\usepackage{adjustbox}  
\usepackage{marvosym}

\newcommand{\tabincell}[2]{\begin{tabular}{@{}#1@{}}#2\end{tabular}} 
\usepackage{tcolorbox}
\usepackage{lipsum}
\usepackage{listings}
\usepackage{courier}
\usepackage{array}
\usepackage{multirow}

\def\paperID{14082} 
\def\confName{ICCV}
\def\confYear{2025}
\usepackage{graphicx}  

\title{UnrealZoo: Enriching Photo-realistic Virtual Worlds for Embodied AI}

\author{
Fangwei Zhong\footnotemark[1]   \textsuperscript{\Letter16} 
Kui Wu\footnotemark[1] \textsuperscript{2} 
Churan Wang\textsuperscript{\textsuperscript{3}}
Hao Chen\textsuperscript{\textsuperscript{4}}
Hai Ci\textsuperscript{\textsuperscript{5}}
Zhoujun Li\textsuperscript{2}
Yizhou Wang\textsuperscript{3}\\
\textsuperscript{1} School of Artificial Intelligence, Beijing Normal University\\
\textsuperscript{2} State Key Laboratory of Complex \& Critical Software Environment, Beihang University\\
\textsuperscript{3} School of Computer Science, Institute for Artificial Intelligence, \\  State Key Laboratory of General Artificial Intelligence, Peking University\\
\textsuperscript{4} City University of Macau  
\textsuperscript{5} National University of Singapore\\
\textsuperscript{6} Beijing Institute for General Artificial Intelligence (BIGAI)\\
{\tt \small fangweizhong@bnu.edu.cn},  {\tt\small wukui@buaa.edu.cn}
}

\begin{document}

\twocolumn[{
\renewcommand\twocolumn[1][]{#1}%

\maketitle

\vspace{-0.6cm}
\includegraphics[width=0.8\linewidth]{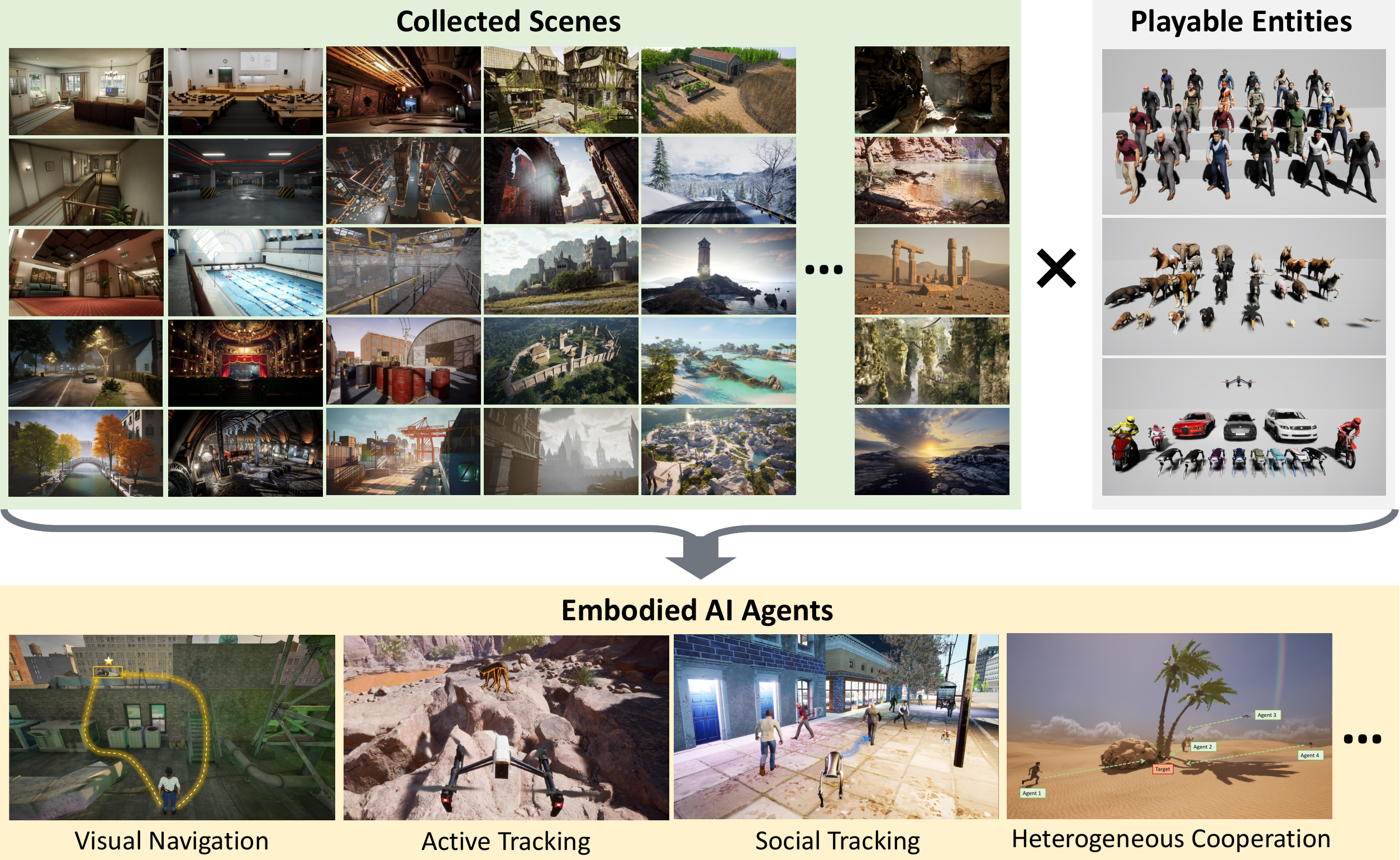}
 \centering
 \vspace{-0.1cm}
\captionof{figure}{
UnrealZoo enriches photo-realistic virtual worlds with diverse scenes and playable entities, enabling training of generalizable embodied AI agents for navigation, tracking, and social interactions.
\vspace{0.25cm}
}
\label{fig:overview}
}]

\renewcommand{\thefootnote}{\fnsymbol{footnote}} 
\footnotetext[1]{Equal contribution, \Letter Corresponding author}
\renewcommand{\thefootnote}{\arabic{footnote}}
\begin{abstract}

We introduce \href{http://unrealzoo.site/}{UnrealZoo}, a collection of over 100 photo-realistic 3D virtual worlds built on Unreal Engine, designed to reflect the complexity and variability of open-world environments. We also provide a rich variety of playable entities, including humans, animals, robots, and vehicles for embodied AI research.
We extend UnrealCV with optimized APIs and tools for data collection, environment augmentation, distributed training, and benchmarking. These improvements achieve significant improvements in the efficiency of rendering and communication, enabling advanced applications such as multi-agent interactions.
Our experimental evaluation across visual navigation and tracking tasks reveals two key insights: 1) environmental diversity provides substantial benefits for developing generalizable reinforcement learning (RL) agents, and 2) current embodied agents face persistent challenges in open-world scenarios, including navigation in unstructured terrain, adaptation to unseen morphologies, and managing latency in the close-loop control systems for interacting in highly dynamic objects.
UnrealZoo thus serves as both a comprehensive testing ground and a pathway toward developing more capable embodied AI systems for real-world deployment.
\end{abstract}

\section{Introduction}

Currently, embodied artificial intelligence (Embodied AI) agents are often \textit{homebodies}, primarily confined to controlled indoor environments and rarely venturing outside to explore the diversity of the open world.  
While several simulators~\citep{kolve2017ai2, CARLA, puig2018virtualhome, li2023behavior,   puig2023habitat} have advanced the field, they often focus on specific scenarios, such as daily activities in homes or autonomous driving on urban roads.
This narrow focus hinders AI agents' adaptability and generalization in open worlds, such as factories, public areas, and landscapes, which are required by several real-world applications.

To bridge this gap, there is a growing demand for simulators that feature diverse open-world environments. 
\emph{First}, the diversity of 3D scenes and morphologies is fundamental to developing spatial intelligence ~\citep{davison2018futuremapping}. This diversity enables agents to actively perceive, reason, plan, and act while tackling various tasks in complex 3D worlds.
\emph{Second}, the complexity of multi-agent interactions is essential for developing social intelligence~\citep{duenez2023social}, such as the theory of mind~\citep{jin-etal-2024-mmtom}, negotiation~\citep{guan2024richelieu}, cooperation~\citep{wang2022tomc}, and competition~\citep{zhong2021distractor}, encouraging agents to behave more like humans.
\emph{Third}, virtual worlds that mimic the challenges in open-world scenarios can evaluate agents efficiently and effectively, identifying limitations and preventing hardware losses from real-world deployment failures~\citep{kadian2020sim2real}.
\emph{Ultimately}, these features will inspire researchers to explore new challenges previously overlooked in other simulators~\citep{duan2022survey}, facilitating seamless integration into real-world applications.

In this work, we introduce UnrealZoo, a comprehensive collection of photo-realistic virtual worlds, based on Unreal Engine and UnrealCV~\citep{qiu2017unrealcv}. UnrealZoo features a diverse range of complex open worlds and playable entities to advance research in embodied AI and related domains. 
This high-quality set includes 100 realistic scenes at varying scales, such as houses, supermarkets, train stations, industrial factories, urban cities, villages, temples, and natural landscapes.
Each environment is expertly designed by artists to mimic realistic lighting, textures, and dynamics, closely resembling real-world experiences.
Our collection also includes diverse playable entities, including humans, animals, robots, drones, motorbikes, and cars. This diversity enables researchers to investigate the generalization of agents across different embodiments or build complex 3D social worlds with numerous heterogeneous agents.
To enhance usability, we further optimize UnrealCV and offer a suite of easy-to-use Python APIs and tools (UnrealCV+), including environment augmentation, demonstration collection, and distributed training/testing. These tools allow for customization and extension of the environments to meet various needs in future applications, ensuring UnrealZoo remains adaptable as the embodied AI agents evolve.

We conduct experiments to demonstrate the applicability of UnrealZoo for embodied AI. First, we benchmark frames per second (FPS) across various commands, highlighting the significant improvement in image rendering and multi-agent interactions with the UnrealCV+ API. We use embodied visual navigation~\citep{thor2017} and tracking~\citep{luo2018end, zhong2021distractor} as two example tasks to benchmark embodied vision agents in complex dynamic environments with moving objects and unstructured maps. We also introduce a set of simple yet effective baseline methods for developing embodied vision agents, including distributed online reinforcement learning algorithms~\citep{mnih2016asynchronous}, offline reinforcement learning algorithms~\citep{kumar2020conservative}, and a reasoning framework for large vision-language models (VLMs). Our evaluations across different settings emphasize the importance of diverse training environments for enhancing agent generalization and robustness, the necessity of low latency in closed-loop control to handle dynamic factors, and the potential of reinforcement learning for training agents to navigate complex scenes.

Our contributions can be summarized as follows:
1) We build UnrealZoo, a collection of 100 high-quality photo-realistic scenes along with a set of playable entities. These cover the most challenging aspects for embodied AI agents in open-world environments.
2) We optimize the communication efficiency of UnrealCV APIs and provide easy-to-use Gym interfaces with a toolkit to meet diverse requirements.
3) We conduct experiments to demonstrate the usability of UnrealZoo, showing the importance of the diversity of the environments to the embodied agents, and analyzing the limitations of the current RL-based and VLM-based agents in the open worlds.

\begin{table*}[t]
\centering
    \caption{The comparison with related photo-realistic virtual worlds for embodied AI.  \textbf{Nav. Sys.} specifies whether the agent in the environment includes an autonomous navigation system. We compare the visual realism across different engines and list the descriptions of the symbols in supplementary materials.}
    \vspace{-0.1cm}
\label{tab:env_comparision}
\begin{tabular}{c}
\begin{minipage}{0.9\textwidth}
\includegraphics[width=\linewidth]{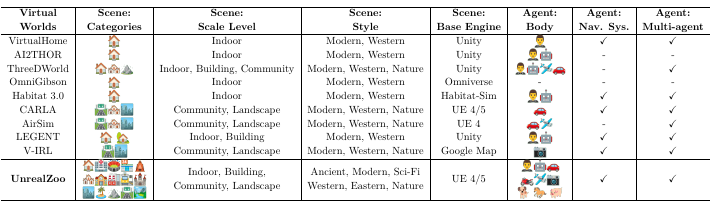}
\end{minipage} 
\end{tabular}
\vspace{-0.5cm}
\end{table*}

\section{Related Works}

\textbf{Realistic Simulators for Embodied AI.}
Realistic simulators are widely adopted in embodied AI research due to their high-quality rendering, cost-effective ground truth generation, and environmental controllability. Specialized 3D simulators have been developed for indoor navigation~\citep{kolve2017ai2, puig2018virtualhome, xia2018gibson, house3D}, robot manipulation~\citep{yu2020meta, ehsani2021manipulathor, chen2023bi}, and autonomous driving~\citep{virtual_kitti, airsim, CARLA}. Recent advances in computer graphics have enabled general-purpose virtual worlds with photo-realistic rendering. ThreeDWorlds (TDW)~\citep{gan2020threedworld} and LEGENT~\citep{cheng2024legent} offer multi-modal platforms based on Unity, but their scenes and playable entities remain limited, with performance degrading in large outdoor environments. V-IRL~\citep{yang2024v} leverages Google Maps' API to simulate agents with real-world street view images, yet lacks physical interaction capabilities due to its static image composition. While the community has begun exploring dynamic environments with social interactions, existing solutions like Habitat 3.0~\citep{puig2023habitat} focus on limited agent interactions in indoor scenes, and HAZARD~\citep{zhou2024hazard} addresses only single-agent simulations in scenarios like fires and floods. In contrast, UnrealZoo offers a comprehensive collection of diverse scenes across different scales, situations, eras, and cultural backgrounds with various playable entities. Leveraging Unreal Engine advancements and optimized UnrealCV, our environment achieves real-time performance in large-scale scenes with multiple agents and photo-realistic rendering. A detailed comparison of photo-realistic simulators is presented in Table~\ref{tab:env_comparision}.

\renewcommand{\arraystretch}{1.1}

\noindent \textbf{Embodied Vision Agents.}
This represents a key frontier in AI research, performing tasks with visual observation, like navigation~\citep{thor2017, gupta2017cognitive, yokoyama2024vlfm, long2024instructnav}, active object tracking~\citep{luo2018end, zhong2018advat, zhong2021distractor, zhong2023rspt, zhong2024empowering}, and other interactive tasks~\citep{chaplot2020learning, weihs2021visual, ci2023proactive, wang2023rearrange}. Their development involves state representation learning~\citep{yadav2023offline, yuan2022pre, gadre2022continuous, yang2023track}, reinforcement learning (RL)~\citep{schulman2017proximal, xu2023drm, ma2023revisiting}, and large vision-language models (VLMs)~\citep{zhang2024navid, zhou2024navgpt}, such as \href{https://openai.com/index/hello-gpt-4o/}{GPT-4o} and SpatialVLM ~\citep{chen2024spatialvlm}.
Despite advances, significant challenges persist. RL methods require extensive trial-and-error interactions and computational resources, often struggling with generalization. VLM-based methods excel at interpreting instructions and images but may lack fine-grained control and adaptability for real-time interactions.
Importantly, previous simulators primarily focus on indoor rooms or urban roads, overlooking critical challenges faced by embodied vision agents in open worlds: unstructured terrain, dynamic factors, perception-control loop costs, and multi-agent social interactions.
Therefore, benchmarking agents in expansive, photo-realistic virtual environments is essential to address real-world challenges. In this work, we collect a subset of environments from UnrealZoo and benchmark embodied visual navigation and tracking agents, revealing limitations in current methods that were previously masked by constrained simulation environments.

\section{UnrealZoo}
UnrealZoo is a collection of photo-realistic, interactive open-world environments with diverse embodied characters, built on Unreal Engine and UnrealCV~\citep{qiu2017unrealcv}.
The environments are sourced from the \emph{Unreal Engine Marketplace}~\footnote{https://www.unrealengine.com/marketplace}, which shares high-quality content from artists, and were accumulated over two years at a cost exceeding $10,000$.
UnrealZoo features a diverse array of scenes with varying sizes and styles. Among them, the largest scene, i.e., Medieval Nature Environment, covers more than $16 km^2$ areas. 
The environments also include a wide range of embodiment, such as human avatars, vehicles, drones, animals, and virtual cameras, all of which can interact with the environment and are equipped with ego-centric sensing systems.
We offer easy-to-use Python APIs based on UnrealCV to facilitate interaction between Python programs and the game engine. 
Note that UnrealCV is optimized for rendering and communication, particularly in large-scale and multi-agent scenarios, namely UnrealCV+.
Additionally, we provide OpenAI Gym interfaces to standardize agent-environment interactions.
The gym-like interface also contains a set of toolkits, e.g., environment augmentation, population control, time dilation, and JSON-style task configurations to help the user customize the environments for various tasks with minimal effort.

\subsection{Scene Collection}

\begin{figure}[t]
    \centering
    \includegraphics[width=1\linewidth]{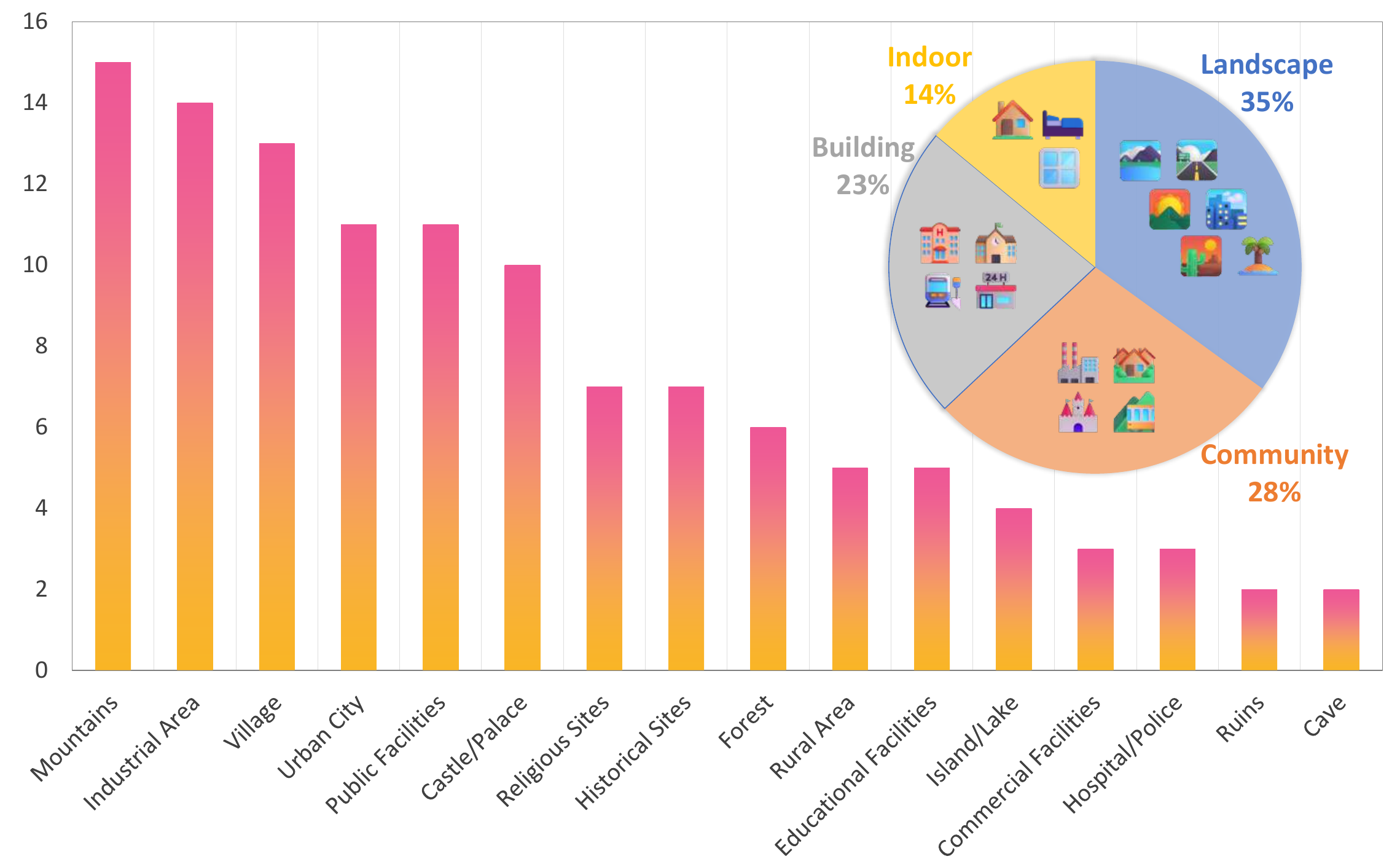}
    \vspace{-0.6cm}
    \caption{The statistical distribution of scene content and scale in UnrealZoo. The bar chart depicts the number of scenarios featuring each type of content, noting that larger scenes may encompass multiple categories. The pie chart classifies these scenes by scale, revealing a predominance of large-scale `Landscape' environments, followed by `Community', `Building', and `Indoor' levels. The distribution reflects the diversity of UnrealZoo and the balanced composition of scenes of different scales. }
    \vspace{-0.5cm}
    \label{fig:stastic_distribution}
\end{figure}

UnrealCV Zoo contains 100 scenes based on Unreal Engine 4 and 5. 
We select the scene based on the public reviews in the marketplace and the difference between the collected scenes, aiming at covering a wide range of styles from ancient to fictional, ensuring diversity. We provide an overview of the environments in the 
\href{https://unrealzoo.notion.site/scene-gallery}{scene gallery}, and the detailed statistical distribution is shown in Figure ~\ref{fig:stastic_distribution}.

We have tagged the collected scenes with several feature labels, allowing researchers to select appropriate scenes for testing or training based on the tags associated with each scene. Our tags cover the following aspects:
\begin{itemize}
    \item \textbf{Scene Categories}: We categorize scenes into three main types: interior, exterior, and both. The interiors include private houses, museums, supermarkets, train stations, factories, gyms, and caves. The exteriors include various outdoor terrains such as ruins, islands, plazas, neighborhoods, and mountains. Additionally, there are 46 scenes that include both interior and exterior elements, offering a blend of architectural elements and natural landscapes, enhancing the versatility and realism of our collection. More details about the distribution of the scenes are shown in Figure~\ref{fig:stastic_distribution}.
    \item \textbf{Scale}: Each scene is labeled according to its scale, and categorized into four levels: indoor, building, community, and landscape. The indoor scale is the smallest, typically encompassing one or multiple interior rooms (up to a complete floor), such as apartments and office interiors. The building scale includes a single building and its immediate surroundings, like a museum, supermarket, or gas station. The community scale covers areas with multiple buildings, such as neighborhoods, villages, castles, or container yards. The landscape scale includes vast natural or man-made areas, or parts of a city or an entire small town, such as mountains, forests, islands, and urban districts.
    Specifically, there are 35 scenes classified as landscape, 28 as community, 23 as building, and 15 as indoor. The largest scene covers 16 square kilometers.
    \item \textbf{Spatial Structure}: 
    We also tag the spatial structure of the scenes, including multi-floor, topological, flat, steep, etc. Such categorization is vital for benchmarking the spatial intelligence of embodied agents. Multi-floor structures, for instance, challenge agents with vertical navigation and require advanced path-planning algorithms. Topological features, such as interconnected pathways, test the agent's ability to understand and traverse complex networks.
    \item \textbf{Dynamics}: The environment's dynamics include simulating factors like weather,  fire, gas, fluid, and interactive objects. These elements enhance the visual and physical diversity of the scene while evaluating agents' adaptability and generalization. Weather variations such as sandstorms, snowfall, and thunderstorms are crucial, as are interactive objects like doors that agents can interact with. These dynamics are vital for an open-world experience.
    \item \textbf{Style}: The scenesen compass diverse cultural and historical styles, including \emph{Asian Temple}, \emph{Western Church}, and \emph{Middle Eastern Street}.  Cultural labels include \emph{Western}, \emph{Asian}, and \emph{Middle Eastern}, while era labels encompass \emph{Medieval}, \emph{Modern}, and \emph{Science Fiction}. Identifying the styles will help us build a new data set to benchmark how social agents adapt to different backgrounds.
\end{itemize}

After categorizing the scenes, we integrate UnrealCV+ (Refer to Section~\ref{sec:unrealcv}) into the UE project and add the controllable player assets (Refer to Section~\ref{sec:body}) to each scene. Due to licensing restrictions, content purchased from the marketplace cannot be open-source, so we package the projects into an executable binary for sharing with the community. These executable binaries will be compatible with various operating systems, including Windows, Linux, and macOS, allowing users to download and run them via the Python interface without needing any knowledge of Unreal Engine, which is primarily built on C++ and Blueprint.

\subsection{Playable Entities}
\label{sec:body}

UnrealZoo includes seven types of playable entities: humans, animals, cars, motorbikes, drones, mobile robots, and flying cameras (See Figure~\ref{fig:overview}). Specifically, it comprises 19 human entities, 27 animal entities (dog, horse, elephant, pig, bird, turtle, etc), 3 cars, 14 quadruped robots, 3 motorcycles, and 1 quad-copter drone. This diversity, with varying affordances like action space and viewpoint, allows us to explore new challenges in embodied AI, such as cross-embodiment generalization and heterogeneous multi-agent interactions.

\begin{figure}[t]
    \centering
    \includegraphics[width=0.99\linewidth]{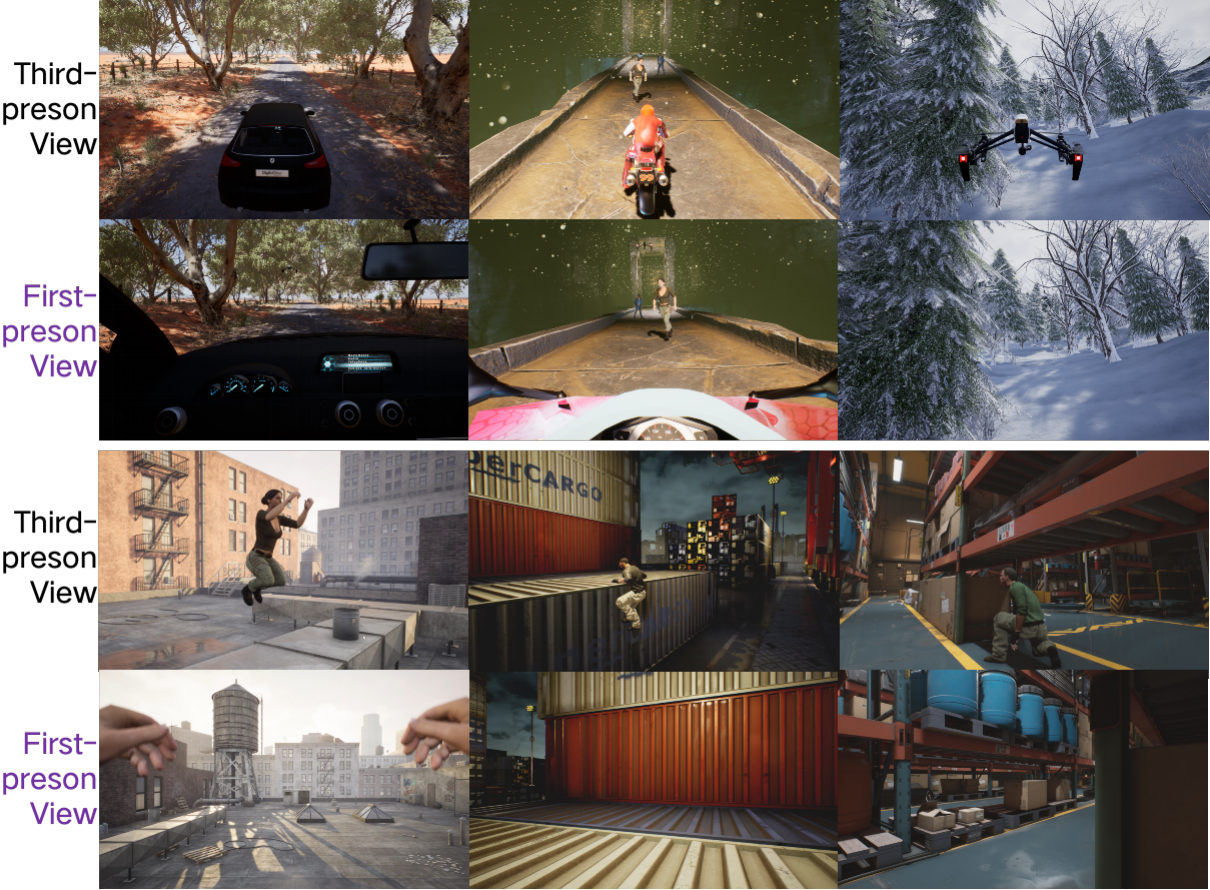} 
    \caption{First-person and third-person view images of different entities in different scenes. Note that camera parameters can be reconfigured by UnrealCV APIs.}
    \vspace{-0.3cm}
    \label{fig:body}
\end{figure}

Each entity includes a skeleton with appropriate meshes and textures, a local motion system, and a navigation system.
We offer a set of callable functions for each entity, enabling users to modify attributes like size, appearance, and camera positions, as well as control movements. 
Each entity can switch between different textures and appearances via UnrealCV API, enhancing visual diversity and adaptability for various scenarios. 
Each entity is equipped with an ego-centric camera, allowing the users to capture various types of image data such as RGB, depth, surface normal, and instance-level segmentation (object mask) from the agent's ego-centric view. Figure~\ref{fig:body} shows examples of the captured first-person view and third-person view images of different entities with varying locomotion.
For multi-agent interaction, the population of the entities in a scene can be easily adjusted using the spawn or destroy functions. 

\textbf{The locomotion system} is built on \href{https://www.unrealengine.com/marketplace/en-US/product/smart-locomotion}{Smart Locomotion}, an animation framework providing natural movement and environmental interactions. Interactions are animation-based rather than physics-based, featuring binary state changes (e.g., opening doors without physical doorknob manipulation) and positional animations (crouching, jumping, climbing), similar to VirtualHome~\citep{puig2018virtualhome}. Interactive objects maintain consistent behavior across all environments and can be spawned at arbitrary locations with customizable visual properties via the API. This approach facilitates research on high-level reasoning and planning without the need for extra implementation of primitive motion control.

\textbf{The navigation system} is built on \href{https://dev.epicgames.com/documentation/en-us/unreal-engine/world-partitioned-navigation-mesh?application_version=5.4}{NavMesh} (a spatial partitioning system that defines walkable surfaces), allowing agents to autonomously navigate with the built-in AI controller.
This includes path-finding and obstacle-avoidance capabilities, ensuring smooth and realistic movement throughout diverse terrains and structures.
For urban-style maps, we segment the roads to distinguish between pedestrian and vehicle pathways. When agents use the navigation system for autonomous control, they will navigate the shortest path based on the priority of the different areas. For example, pedestrians and animals will prioritize walking on sidewalks, while vehicles and motorcycles will prioritize driving on roadways. An example of the navigation area is shown in 
 Supplementary Materials.

\subsection{Programming Interface}

\begin{figure}[t]
    \centering
    \includegraphics[width=1.0\linewidth]{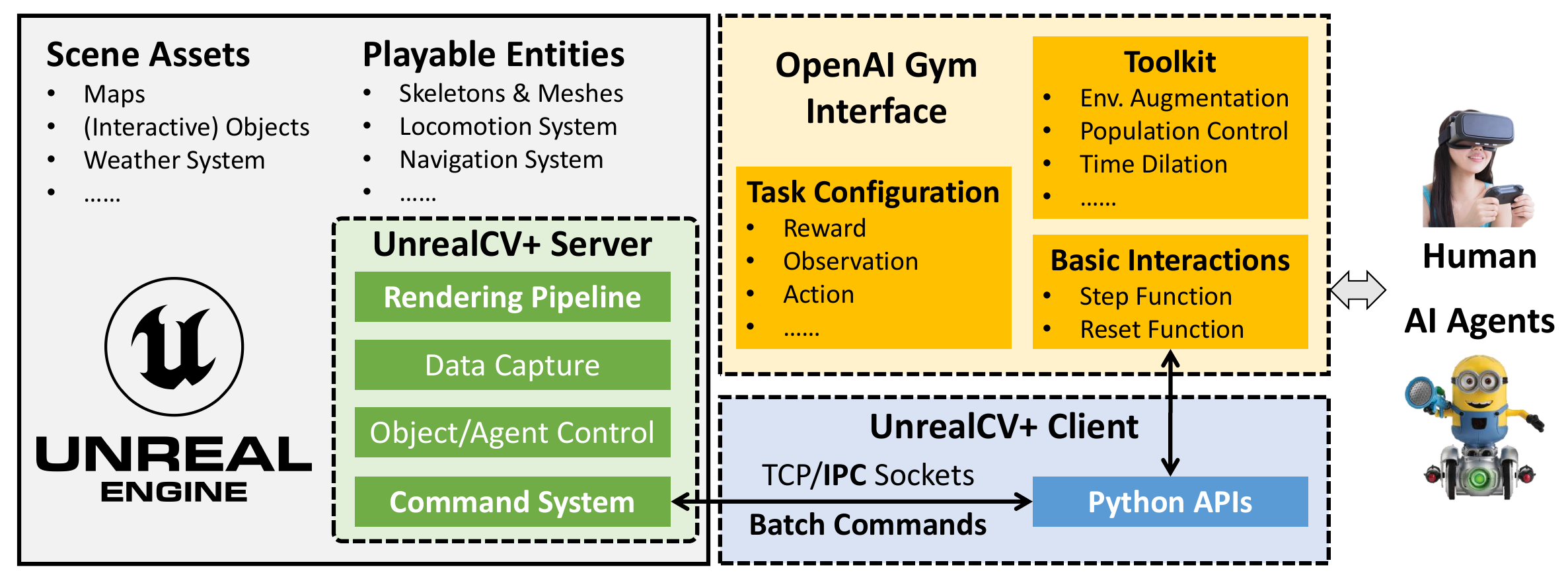}
    \caption{The detailed architecture of UnrealZoo. The Gray box indicates the UE binary, containing the scenes, playable entities, and UnrealCV+ server. We have bolded the names of the optimized or new modules in UnrealCV+ Server and Client. We provide OpenAI Gym Interfaces for agent-environment interaction. Our gym interface supports customizing the task in a configuration file and contains a toolkit with a set of gym wrappers for environment augmentation, population control, and etc.}
    \vspace{-0.4cm}
    \label{fig:architecture}
\end{figure}

\label{sec:unrealcv}
We provide UnrealCV+ as the fundamental interface on Python to capture data and control the entities and scenes, and provide an OpenAI Gym interface~\footnote{https://github.com/UnrealZoo/unrealzoo-gym} for general agent-environment interactions. The architectures of the programming interfaces are shown in Figure~\ref{fig:architecture}.

\begin{table*}[tb]
\centering
\vspace{-0.2cm}
\caption{Comparison of FPS in Unreal Engine 4.27 with UnrealCV and UnrealCV+. Results represent average performance across 6 typical environments ranging from small-scale scenes ($2,440 m^2$) to expansive landscapes ($16km^2$). Benchmarks were conducted on a system with an Nvidia RTX 4090 GPU, Intel i7-14700k CPU, and Windows OS. Performance metrics were collected by executing each Python function 1,000 times and measuring execution duration. Multi-agent interaction testing utilized the gym interface, with all rendered images at 640×480 resolution. See the supplementary material for the detailed results.}
 \resizebox{0.8\linewidth}{!}{
\begin{tabular}{l|c c c c|c c c}
\hline
 & \multicolumn{4}{c|}{Image Capture} & \multicolumn{3}{c}{Multi-agent Interaction} \\ 
    & Color & Object Mask & Surface Normal & Depth & N=2 & N=6 & N=10 \\ \hline
UnrealCV  & 74 &70 & 109&52 & 35  & 13 & 8 \\ \hline
UnrealCV+  & $83(\uparrow12\%)$ & $154(\uparrow120\%)$ & $131(\uparrow20\%)$ & $97(\uparrow86\%)$ & $54(\uparrow54\%)$ & $25(\uparrow92\%)$ & $16(\uparrow100\%)$ \\ \hline
\end{tabular}
}
\label{tab:fps_image}
\vspace{-0.3cm}
\end{table*}

\textbf{UnrealCV+} is our improved version of the UnrealCV~\citep{qiu2017unrealcv} for high-throughput interactions. As the original version of UrnealCV primarily focuses on generating synthesis data for computer vision, the frame rates per second (FPS) are not optimized for real-time interactions.
We optimize the rendering pipelines in the UnrealCV server and the communication protocols between the server and the client to improve the FPS. Specifically, we enable parallel processing while rendering object masks and depth images, which can significantly improve the FPS in large-scale scenes.
For multi-agent interactions, we further introduce the batch commands protocol. In this protocol, the client can simultaneously send a batch of commands to the server, processing all the received commands and returning a batch of results. In this way, we can reduce the time spent on server-client communication. Since reinforcement learning requires extensive trial-and-error interactions, often running multiple environments on a computer, we introduce Inter-process communication (IPC) sockets instead of TCP sockets to improve the stability of the server-client communication under high loads. We benchmark the FPS performance in Table~\ref{tab:fps_image}. 
To enhance user-friendliness, we have developed high-level Python APIs that are built upon the command systems of UnrealCV. These APIs encapsulate all the request commands and their corresponding data decoders into a callable Python function. This approach significantly simplifies the process for beginners, allowing them to interact with and customize the environment using UnrealCV+.

\textbf{Gym Interface} is used to define the interactive tasks and standardize the agent-environment interaction, following \href{https://github.com/zfw1226/gym-unrealcv}{Gym-UnrealCV}.
Even though there are a lot of tasks for agents, they usually share common interaction protocols, i.e., the agent gets observations from the environment and returns actions. The main difference across different tasks is usually the reward functions, the modality of the observation, and the available actions. Hence, we define the basic interaction functions for general usage and list the task-specific configurations, e.g., scene name, and reward function, in a JSON File. In this way, when adding new UE scenes, the users only need to set the parameters in the JSON files. Moreover, we provide a toolkit with a set of gym wrappers for training and testing the agents, such as environment augmentation that has been in previous work for training generalizable agents~\citep{luo2018end, luo2019pami}, population control to adjust the number of agents in the scene, and time dilation to adjust the control frequency in dynamic scenes. In Section~\ref{sec:social_tracking}, we demonstrate an example usage of the toolkit to analyze the robustness of social tracking agents to the population of crowds and the impact of the control frequency in such dynamic scenes. We also provide a launch tool to enable the user to run multiple environments with specific GPU IDs within a computer, which is useful for distributed online reinforcement learning. 

\vspace{-0.3cm}
\section{Experiments}
\vspace{-0.3cm}

In this section, we conduct experiments involving a series of challenging tasks on UnrealZoo to show its unique features regarding usability, diversity, efficiency, and extensibility. We begin with a visual navigation task to indicate the suitability of UnrealZoo for distributed online reinforcement learning. This task also highlights how the complex spatial structures within our environments introduce new challenges that the off-the-shelf methods struggle to address effectively. Next, we evaluate agents through active visual tracking to validate the necessity of scaling environmental diversity for improving cross-scene generalization. Finally, our social tracking experiments illustrate the exceptional flexibility of UnrealZoo. By enabling easy modification of simulation control frequencies, population scale, and agent morphologies, UnrealZoo reveals previously overlooked challenges in high-dynamic environments and cross-body generalization.

\vspace{-0.2cm}
\subsection{Emergent Challenges in Visual Navigation} 
\vspace{-0.2cm}
Taking the visual navigation task as an example, a fundamental skill for embodied agents, we demonstrate the usability of our environments for various approaches, including distributed online reinforcement learning and large vision-language models. Additionally, we highlight the emergent challenges posed by unstructured open-world environments. 
While traditional approaches often operate within well-defined 3D spatial structures, our environments require agents to navigate through complex, unconstrained terrains.
Therefore, agents must execute diverse locomotion skills, including running, climbing, jumping, and crouching, to overcome various environmental obstacles while pursuing target objects.
This setting demands sophisticated 3D spatial reasoning capabilities and action selection mechanisms for effective real-time decision-making.
The details of the task setting, implementation, and training curve are introduced in the supplementary materials.

\textbf{Evaluation Metrics.}
We employ two key metrics to evaluate visual navigation agents: 1) Average Episode Length (EL), representing the average number of steps per episode over 50 episodes. 2) Success Rate (SR), measuring the percentage of episodes the agent successfully navigates to the target object out of 50 total episodes, which represents the navigation capability in the wild environment. 3) Success weighted by Path Length (SPL)~\citep{anderson2018evaluation}, a standard metric that accounts for both success and path efficiency.

\textbf{Baselines for Navigation.} We build simple baselines to demonstrate the applicability of our environments for training reinforcement learning agents and benchmark the agents based on pre-trained large models.
\textbf{1) Online RL}: We train RL-based navigation agents independently in the Roof and Factory environments using a distributed online reinforcement learning (RL) approach, e.g., A3C~\citep{mnih2016asynchronous}. The model takes the first-person view segmentation mask and the relative position between the agent and target as input, and outputs discrete actions to navigate. This setup allows the agent to iteratively refine navigation strategies through continuous environmental interaction. 
\textbf{2) GPT-4o}: We leverage the multimodal reasoning capabilities of GPT-4o for navigation model.
Following a structured prompt template, GPT-4o performs reasoning to select appropriate actions from the action space to guide the agent toward the target location.
\textbf{3) Human}: As a reference, we conduct experiments where human players control agents via keyboard inputs, analogous to first-person video game interactions. The human players use the same action space available to other baselines.

\textbf{Results.}
In Table~\ref{tab:nav_res}, we report the performances of different methods in two unstructured scenes. The RL-based agent performs moderately well, achieving better results in the simpler environment (\emph{Industrial Area}) compared to the \emph{Roof}, where the target object is located on different levels of stairs. The GPT-4o agent struggles in both scenarios. This implies that the GPT-4o performs poorly in complex 3D scene reasoning. As a reference, the human player completes both tasks with the fewest steps and a 1.00 success rate, underscoring the significant performance gap between current embodied AI agents and humans, indicating substantial room for improvement to navigate in such complex, open-world environments.

\begin{table}
\centering
\caption{The results (EL/SR/SPL) of visual navigation agents in two unstructured terrains.}
\begin{adjustbox}{width=\columnwidth}
\begin{tabular}{c|ccc}
\hline
 & \bf Online RL & \bf GPT-4o & \bf Human \\  
\hline 
Roof & 1660/0.32/0.10 & 2000/0.00/0.00 & 515/1.00/0.97 \\  
Industrial Area & 261/0.52/0.31 & 369/0.20/0.08 & 158/1.00/0.98 \\  
\hline
\end{tabular}
\end{adjustbox}
\label{tab:nav_res}
\end{table}

\begin{figure}[t]
    \centering
    \includegraphics[width=0.99\linewidth]{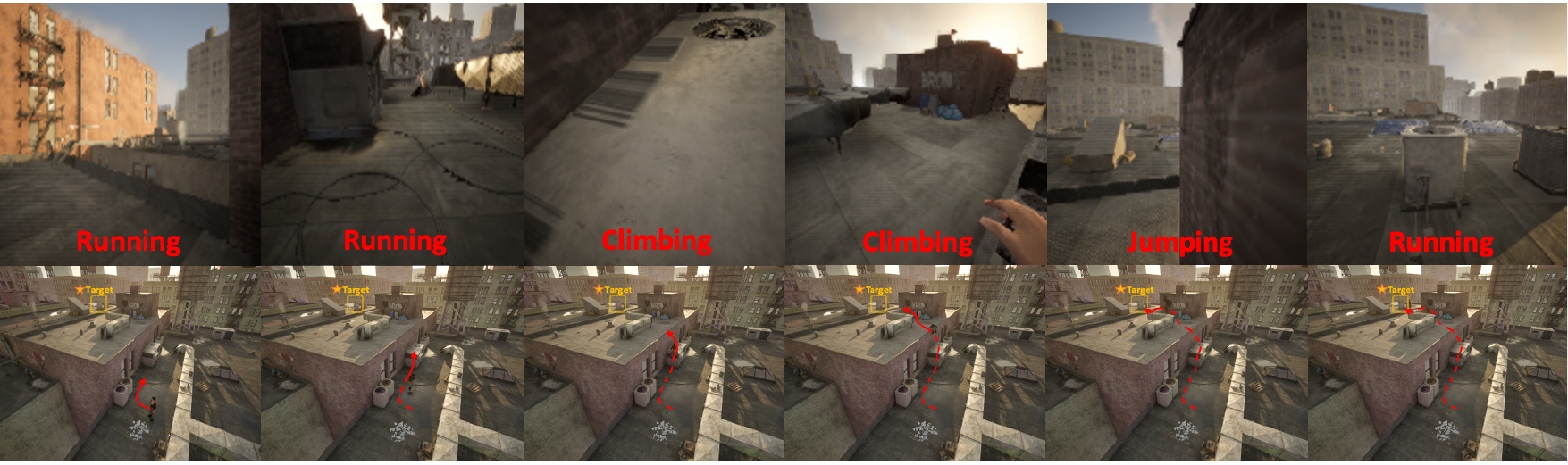}
    \caption{An exemplar sequence from the embodied navigation agent in the \emph{Roof}. The RL agent learned to climb on a box and wall and jump over an obstacle to reach the goal in a short path.}
    \label{fig:eval_example}
    \vspace{-0.5cm}
\end{figure}

\subsection{Scaling the Diversity for Active Visual Tracking}
In this section, we aim to analyze the effectiveness of the diversity of the training data on the generalization of the active visual tracking.
We collect 3 sets of demonstrations from different numbers of environments in UnrealZoo and train the tracking policy via offline reinforcement learning~\cite{zhong2024empowering}.
We evaluate the tracking performance across 16 unseen environments covering four categories: \emph{Interior Scenes}, \emph{Palaces}, \emph{Wilds}, and \emph{Modern Scenes}. 
Each category contains 4 individual environments.
The implementation details are introduced in the supplementary materials.

\textbf{Evaluation Metrics.}
Our evaluation employs three key metrics: (1) Average Episodic Return (ER): calculates the mean episodic return over 50 episodes, providing insights into overall tracking performance; (2) Average Episode Length (EL): represents the average number of steps per episode, which reflects long-term tracking effectiveness; and (3) Success Rate (SR): measures the percentage of episodes that complete 500 steps out of 50 total episodes. 

\textbf{Baselines for Active Visual Tracking.}
1) \textit{PID}: A classical control approach that uses a PID controller to adjust agent actions by maximizing IoU between the detected target bounding box and an expected position.
2) \textit{RL}: We adapt the offline reinforcement learning implementation from \citet{zhong2024empowering}, using their network architecture while collecting three distinct offline datasets (100k steps each) with varying environmental diversity: single environment (\textit{1 Env.}), two environments (\textit{2 Envs.}), and eight environments (\textit{8 Envs.}). Notably, \emph{FlexibleRoom} represents an abstract environment where objects appear as geometric shapes with randomized patterns used in the three settings.
3) \textit{OpenVLA}:  We fine-tune OpenVLA \citep{kim2024openvla}, an end-to-end large model originally designed for robotic manipulation with strong generalization capabilities, on our most diverse dataset (\textit{8 Envs.}) to adapt it for tracking tasks.
4) \textit{GPT-4o}: We implement a VLM-based agent using GPT-4o to generate actions directly from observed images. We designed a prompt template to standardize outputs into predefined action commands (moving forward/backward, turning left/right, or maintaining position), which are then mapped to appropriate velocities. Despite GPT-4o's capabilities, we observed poor alignment performance and significant latency issues when processing raw image inputs, limiting its effectiveness for real-time tracking.

\begin{figure}[t]
    \centering
    \includegraphics[width=\linewidth]{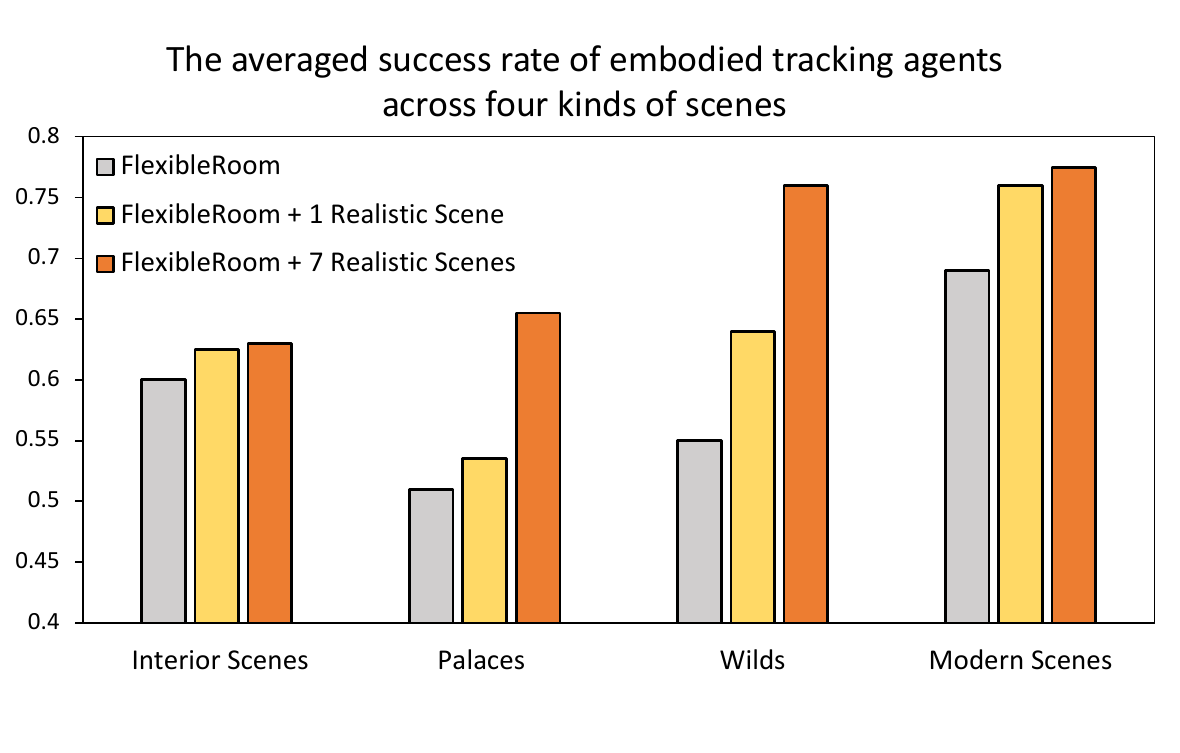}
    \vspace{-0.8cm}
    \caption{Average success rates across four environment categories: Interior Scenes, Palaces, Wilds, and Modern Urbanity.
    Results compare Offline RL agents trained on three offline dataset settings with increasing environmental diversity (1 Env., 2 Envs., 8 Envs.). Performance consistently improves with dataset diversity, demonstrating enhanced generalization capabilities. Notably, environments with complex spatial structures (Interior Scenes and Palaces) present greater challenges for navigation and obstacle avoidance, resulting in lower success rates.
    }
    \vspace{-0.4cm}
    \label{fig:eval_SR}
\end{figure}

\textbf{Result Analysis.}
We first evaluate the performance of agents trained with offline datasets collected from varying numbers of environments (1 Env., 2 Envs., 8 Envs.) across \textbf{16 unseen environments}
To visualize the performance change of different settings within various scene categories, we calculate the average success rate (SR) of each agent in four categories, and the results are shown in Figure~\ref{fig:eval_SR}. Then we compared the baseline methods' performance, as shown in Table~\ref{tab:eval_robust} (0D). The results reveal a clear trend: \textbf{as the number of environments used for training increases, agent long-term tracking performance generally improves across all categories.} In the Wilds, a significant increase in success rate is observed with the 8 Envs. dataset, which involves the highest diversity of environments. This demonstrates that diverse environmental exposure plays a crucial role in improving the agent's generalization capabilities in more complex, open-world environments. The lower success rate in the 1 Env. highlights the limitations of training solely in abstract settings like the FlexibleRoom. Similarly, in the Palace, the success rate improves notably from 1 Env. to 8 Envs., suggesting that training with a broader range of environments helps the agent better adapt to intricate 3D spatial structures.

\subsection{Benchmarking Agents at Variant Settings}
\label{sec:social_tracking}
We augment the environment employed for active tracking by seamlessly integrating the Gym toolkit into diverse settings.
We use the \textit{population control wrapper} to create social tracking scenarios. In this setting, the agent is required to follow the target through crowds.
We can also adjust the morphology of the trackers for evaluating the cross-embodiment generalization.
Moreover, we employ the \textit{time dilation wrapper} to simulate a variety of control frequencies. This allow us to analyze the impact of the latency within the perception-control loop on tracking performance.

\textbf{Robustness to Active Distractions.}
A key challenge in active visual tracking is handling visual distractions and dynamic obstacles, which is crucial for real-world deployment in crowds. We evaluated agents in four representative environments: \underline{Storage House}, \underline{Desert Ruins}, \underline{TerrainDemo}, and \underline{SuburbNeighborhoodDay}, with varying numbers of human distractors (0D, 4D, 10D) and compared RL-based methods (trained in 8 environments, denoted as RL 8 Envs.) against PID, OpenVLA, and GPT-4o.
Table~\ref{tab:eval_robust} shows that RL-based methods maintain stable success rates, RL 8 Envs. achieving 0.76 SR in 0D, 0.68 in 4D and 0.56 in 10D, benefiting from diverse training data. In contrast, PID, OpenVLA, and GPT-4o struggle, showing significantly lower returns and success rates.
PID suffers from delayed reactions when the target is occluded, leading to failures. OpenVLA tracks effectively at short distances but loses accuracy over time due to accumulated errors. Both OpenVLA and GPT-4o are highly prone to distractions, often tracking the wrong target.
These results highlight the limitations of large models in dynamic environments with active distractions. 

\textbf{Cross-Embodiment Generalization.} To investigate whether policies can effectively transfer to agents with different morphologies, we apply the agents (RL and OpenVLA) trained on a human character to a quadrupedal robot dog. This exploration is motivated by the practical necessity in real-world applications where a single policy might need to operate across various embodiments with different perceptual perspectives and movement capabilities. As shown in Table ~\ref{tab:eval_robust}, the performance decreases notably, particularly the success rate (from 0.68 to 0.48 with 4D and from 0.56 to 0.40 with 10D for RL 8 Envs.), indicating that the community should pay more attention to the cross-embodiment generalization for real-world deployment.

\textbf{The Impact of Control Frequency.} 
In high-dynamic scenes, the latency of the perception-control loop plays a critical role in managing multiple dynamic factors effectively. To verify this,
We employ the \textit{time dilation} wrapper to simulate different control frequencies during deployment. As shown in Table~\ref{tab:frequency_control}, when the rate drops below 10 FPS, performance significantly declines.
Our results show that RL-based agents achieve superior performance in social tracking scenarios when operating at higher control frequencies. 
These findings emphasize the necessity of developing efficient models for embodied agents to successfully execute tasks in dynamic social worlds.

\begin{table} \centering 
\small
\caption{The impact of control frequency on tracking performance. We evaluate the RL agent (1 Env.) in the \underline{FlexibleRoom} using the time dilation wrapper to simulate varying control frequencies.} 
\begin{tabular}{c|c|c|c|c} 
\hline 
& 3 FPS & 10 FPS & 30 FPS & w/o FPS Control \\ \hline
ER & 184 & 303 & \textbf{368} & 275 \\ 
EL & 377 & 449 & \textbf{482} & 425 \\
SR & 0.34 & 0.62 & \textbf{0.92} & 0.74 \\ \hline 
\end{tabular} 
\label{tab:frequency_control} 
\vspace{-0.3cm}
\end{table}

\begin{table}[tb]
    \centering
    \caption{ We evaluate different methods in four representative environments: \underline{Storage House}, \underline{Desert Ruins}, \underline{TerrainDemo}, and \underline{SuburbNeighborhoodDay}, with varying numbers of distractors (0D, 4D, 10D). Each cell presents the \textbf{averaged values across four environments} for three metrics: Episodic Return (ER), Episode Length (EL), and Success Rate (SR). OpenVLA and RL are trained on the dataset collected from 8 environments.}
\begin{adjustbox}{width=\columnwidth}

\begin{tabular}{c|ccc}
\hline
      Method & \tabincell{c}{0D} & \tabincell{c}{4D}  &  \tabincell{c}{10D} \\ \hline
PID &113/298/0.34&107/295/0.30& 78/105/0.14\\
GPT-4o& 2/228/0.28& -102/264/0.16	&-80/240/0.10 \\
OpenVLA&55/323/0.10 &-39/149/0.02 & -30/156/0.00 \\
RL &\textbf{321/468/0.76} &\textbf{280/437/0.68} &\textbf{187/402/0.56} \\
\hline
PID (Robot dog) &152/245/0.30&149/237/0.22 & 100/102/0.16\\ 
GPT-4o (Robot dog) &-5/263/0.20&-98/272/0.16 & -59/223/0.04\\
OpenVLA(Robot dog) &42/301/0.08&-37/142/0.00&-13/117/0.00\\
RL (Robot dog)& 279/433/0.60&220/409/0.48  & 143/367/0.40\\ 

\hline
\end{tabular}
\end{adjustbox}

    \vspace{-0.5cm}
    \label{tab:eval_robust}
\end{table}

\section{Conclusions}
In conclusion, UnrealZoo addresses critical gaps in embodied AI research by providing unprecedented environmental diversity across 100 scenes and 67 embodiment options.  
By offering easy-to-use Python APIs and toolkits, we support efficient environmental interactions, distributed training, and task customization in UnrealZoo.
The experiments on UnrealZoo reveal the importance of environment diversity for agent generalization capabilities, the necessity of low-latency control in dynamic scenes, and key challenges faced by current methods for embodied agents in open worlds. These findings highlight a clear path forward for developing more robust embodied AI systems capable of navigating the complexity of real-world deployment.

Looking forward, UnrealZoo will drive research in several key areas: 1) spatial intelligence development in complex 3D environments; 2) social intelligence research through multi-agent interactions; 3) safe pre-deployment evaluation and limitation identification; 4) cross-embodiment learning and skill transfer; and 5) adaptation to dynamic environments. As we continue to enrich virtual worlds and interaction capabilities, UnrealZoo will play an increasingly vital role in advancing agents from simulation to reality, ultimately fostering harmonious human-AI coexistence.

\section*{Acknowledgments}
This work was supported by the National Science and Technology Major Project (2022ZD0114904), NSFC-6247070125, NSFC-62406010, the 	
Funding Scheme for Research and Innovation of FDCT (Grant No. 0021/2025/ITP1), the Fundamental Research Funds for the Central Universities, and Qualcomm University Research Grant. We express our gratitude to Weichao Qiu for his invaluable support in optimizing UnrealCV and for the insightful discussions. We are also thankful to Tingyun Yan for providing the early version of the human entity and to Jingzhe Lin for his assistance in producing demo videos. Our appreciation extends to the active developers in the open-source community for ensuring UnrealCV's compatibility with the latest Unreal Engine versions, such as UE 5.4. Furthermore, we thank the contributors in the Unreal Engine Marketplace for offering high-quality content and plugins, which have been instrumental in the development of UnrealZoo.

{
    \small
    \bibliographystyle{ieeenat_fullname}
    \bibliography{main}
}

\clearpage
\appendix
\tableofcontents
\input{appendix}

\end{document}

%% file: math_commands.tex
\usepackage{amsmath,amsfonts,bm}

\def\eqref#1{equation~\ref{#1}}

\def\1{\bm{1}}

\DeclareMathAlphabet{\mathsfit}{\encodingdefault}{\sfdefault}{m}{sl}
\SetMathAlphabet{\mathsfit}{bold}{\encodingdefault}{\sfdefault}{bx}{n}



%% file: appendix.tex
\clearpage
\section{UE Environments}
\label{app:env}

\subsection{Comparison with other Simulators}
\begin{figure*}[t]
    \centering
    \includegraphics[width=0.9\linewidth]{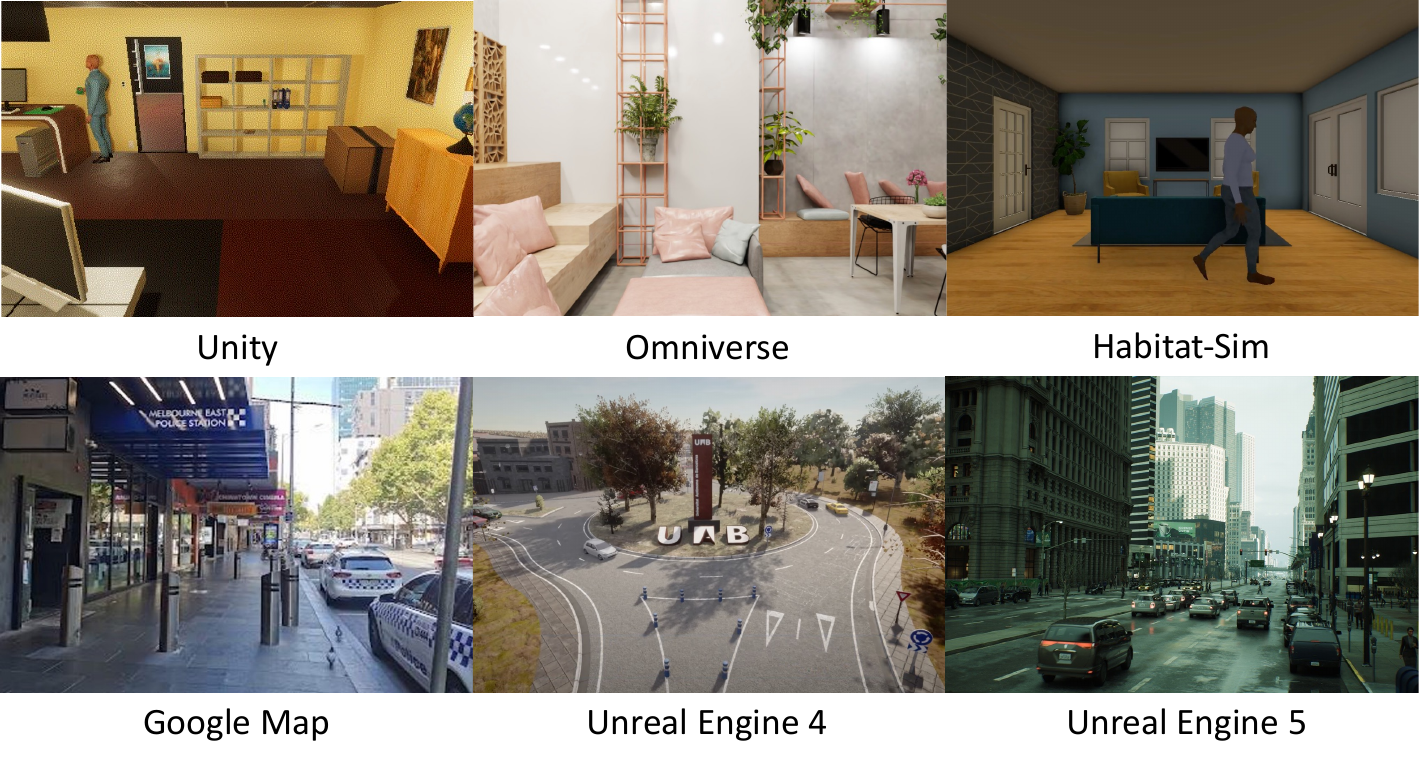}
    \vspace{-0.3cm}
    \caption{Comparison of the visual realism of different engines: we show the snapshots captured from different engines to compare the photo-realism of different environments for an intuitive feeling. Note that Google Maps captures and reconstructs the images from the real world, but can not simulate the dynamics of the scenes and interactions between agents and objects.}
    \label{fig: visual_realism}
\end{figure*}

To better explain Table~\ref{tab:env_comparision}, we list the description of each symbol about the scene types and playable entities in Table~\ref{tab:symbol}. Since photorealism mainly relies on the engine used, we visualize the snapshots rendered by different engines in Figure~\ref{fig: visual_realism}. Note that Google Maps are images captured in the real world, but can not simulate the dynamics of the scenes and interactions between objects. By utilizing advanced rendering and physics engines, Unreal Engine simulates large-scale photorealistic environments that are not only visually appealing but also capable of complex interactions between agents and objects. So we choose to build environments on Unreal Engine.

\begin{table*}[h]
    \centering
     \caption{We evaluate the simulation speed on an Nvidia GTX 4090 GPU, Intel i7-14700K CPU, and Windows OS by repeatedly calling UnrealCV+ API function 1000 times and recording the execution time. Multi-agent interaction performance is measured using the gym interface. The image resolution is set to 640×480. The table presents the FPS measurement results across environments of varying scales.}
    \renewcommand{\arraystretch}{1.2}
    \resizebox{\textwidth}{!}{ 
    \begin{tabular}{lccccccc}
        \toprule
        \textbf{Env} & \textbf{Color Image} & \textbf{Object Mask} & \textbf{Surface Normal} & \textbf{Depth Image} & \textbf{2 Agents Interaction} & \textbf{6 Agents Interaction} & \textbf{10 Agents Interaction} \\
        \midrule
        \textbf{FlexibleRoom} (71 objects, 2440m$^2$) & 85  & 164  & 137  & 100  & 94  & 40  & 27  \\
        \textbf{BrassGarden} (467 objects, 9900m$^2$) & 107  & 214  & 173  & 123  & 102  & 48  & 31  \\
        \textbf{Supermarket} (2839 objects, 11700m$^2$) & 99  & 173  & 167  & 117  & 70  & 33  & 19  \\
        \textbf{SuburbNeighborhoodDay} (2469 objects, 23100m$^2$) & 79  & 139  & 112  & 92  & 53  & 27  & 17  \\
        \textbf{GreekIsland} (3174 objects, 448800m$^2$) & 82  & 167  & 124  & 103  & 52  & 23  & 16  \\
        \textbf{MedievalNatureEnvironment} (8534 objects, 16km$^2$) & 70  & 112  & 97  & 74  & 29  & 14  & 10  \\
        \bottomrule
    \end{tabular}
    }
   
    \label{tab:fps_comparison}
\end{table*}

\begin{table*}[t]
\centering
    \caption{The comparison with related photo-realistic virtual worlds for embodied AI.}
    \vspace{-0.1cm}
\label{fig:env_comparision}
\begin{tabular}{c}
\begin{minipage}{1\textwidth}
\includegraphics[width=\linewidth]{image/Compare_sim.pdf}
\end{minipage} 
\end{tabular}
\vspace{-0.5cm}
\end{table*}

\begin{table*}[h]
\centering
\caption{The description of symbols used in Table~\ref{tab:env_comparision}.}
\vspace{-0.2cm}
\label{tab:symbol}
\includegraphics[width=0.9\linewidth]{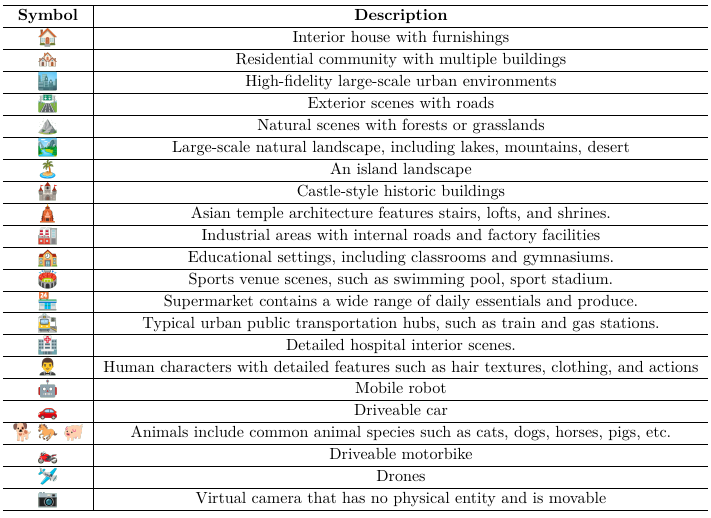}
\end{table*}

\subsection{Environments used in Visual Navigation}
\begin{figure}[t]
    \centering
    \includegraphics[width=0.9\linewidth]{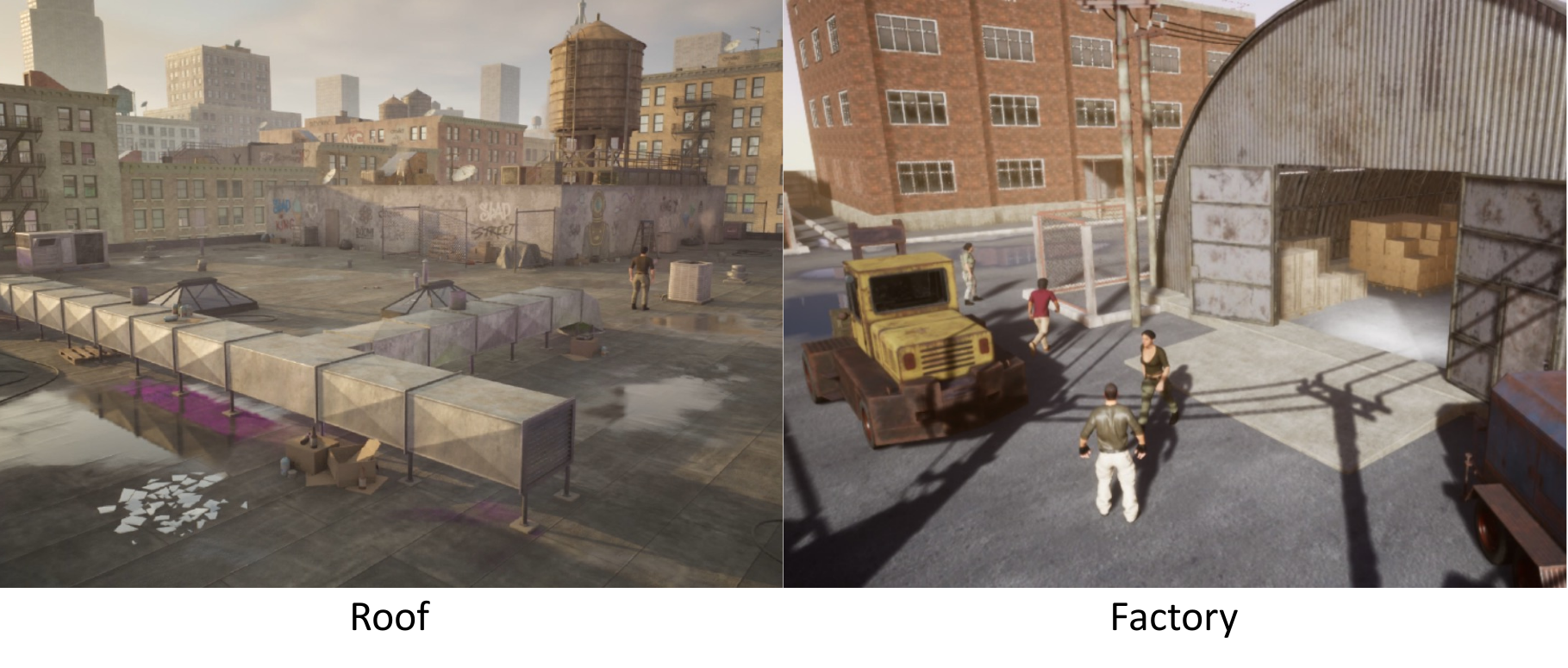}
    \caption{ Two photo-realistic environments used for visual navigation. }
    \label{fig:Nav_env}
\end{figure}

We carefully selected two photo-realistic environments (\textbf{Roof} and \textbf{Factory}) for training and evaluating navigation in the wild, shown in Figure~\ref{fig:Nav_env}. The Roof environment features multiple levels connected by staircases and large pipelines scattered on the ground, providing an ideal setting for the agent to learn complex action combinations for transitioning between levels, such as jumping, climbing, and navigating around obstacles. The Factory environment, on the other hand, is characterized by compact boxes and narrow pathways, challenging the agent to determine the appropriate moments to jump over obstacles or crouch to navigate under them. These two environments offer diverse spatial structures, enabling agents to develop an understanding of multi-level transitions and precise obstacle avoidance.

\subsection{Environments used in Active Visual Tracking}
\label{app:env_track}
\begin{figure*}[t]
    \centering
    \includegraphics[width=1\linewidth]{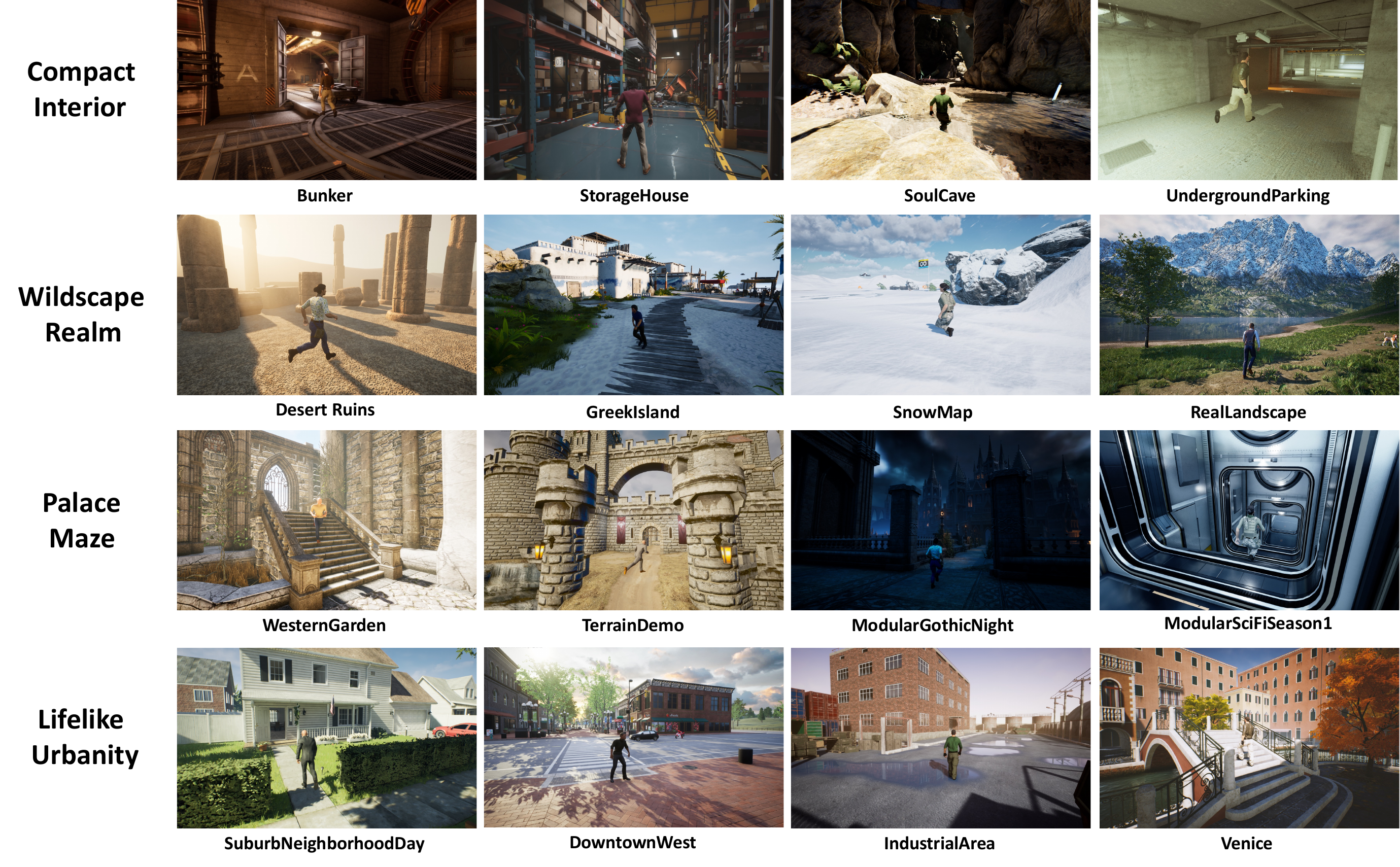}
    \caption{The snapshots of 16 environments used for testing active visual tracking agents. The text on the left indicates the category corresponding to that line of environment.}
    \label{fig:eval_env}
\end{figure*}

For training agents via offline reinforcement learning, we selected 8 distinct environments to collect demonstrations, as is shown in Figure~\ref{fig:data_distribution}. 
To comprehensively evaluate the generalization of the active visual tracking agents, we selected \textbf{16} distinct environments, categorized into Interior Scenes, Palaces, Wilds, and Modern Scenes. Each category presents unique challenges: 1) \textbf{Interior Scenes} feature complex indoor structures with frequent obstacles; 2) \textbf{Palaces} include multi-level structures and narrow pathways; 3) \textbf{Wilds} encompass irregular terrain and varying illumination; 4) \textbf{Modern Scenes} offer high-fidelity, real-world scenarios with modern buildings and objects. These diverse environments facilitate a thorough assessment of the agent’s generalization capabilities across varying complexities. The snapshot of each environment is shown in Figure ~\ref{fig:eval_env}.

\subsection{Navigation Mesh}
\begin{figure}[t]
    \centering
    \includegraphics[width=0.8\linewidth]{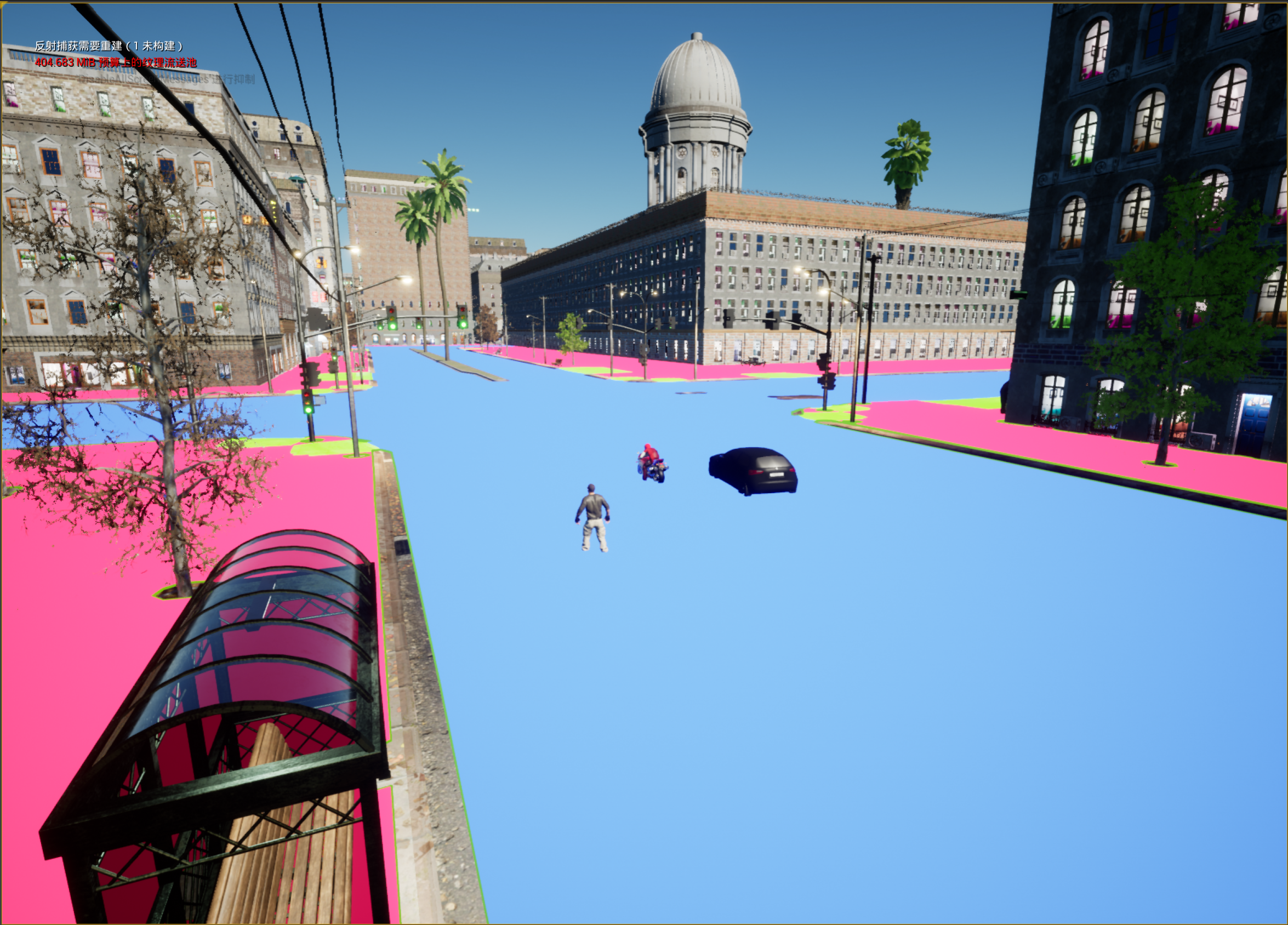}
    \caption{An example of the NavMesh with semantic segmentation. The human character will prioritize using the pink area for pedestrian navigation tasks, while the vehicles will use the blue area.}
    \label{fig:nav_mesh}
\end{figure}

 Based on \href{https://dev.epicgames.com/documentation/en-us/unreal-engine/world-partitioned-navigation-mesh?application_version=5.4}{NavMesh}, we build an internal navigation system, allowing agents to autonomously navigate with the built-in AI controller in the Unreal Engine. This includes path-finding and obstacle-avoidance capabilities, ensuring smooth and realistic movement throughout diverse terrains and structures. Moreover, in our City style map, we manually construct road segmentation, and we manually segment the roads to distinguish between pedestrian and vehicle pathways. When agents use the navigation system for autonomous control, they will navigate the shortest path based on the priority of the different areas. Figure~\ref{fig:nav_mesh} shows an example of the rendered semantic segmentation for NavMesh in an urban city.
 
\section{Exemplar Tasks}

\subsection{Visual Navigation}
\label{app:navigation_task}
In this task, the agent is initialized at a random location in the environment at the beginning of each episode, while the target object's location and category remain fixed throughout. The agent must rely on its first-person view observations and the relative spatial position of the target as input. The ultimate objective is to locate the target object within 2000 steps. Success is defined by the agent reducing the relative distance to less than 3 meters and aligning its orientation such that the relative rotation between the target and the agent is smaller than 30 degrees (in the front of the agent). This setup challenges the agent to optimize its movements and decision-making while adapting to the randomized starting conditions and dynamic environment. All methods in the task share the same discrete action space to control the movement, consisting of moving forward (+1 meter/s), moving backward (-1 meter/s), turning left (-15 degrees/s), turning right (+15 degrees/s), jumping (two continuous jumping actions trigger the climbing action), crouching, and holding position. This action space enables the agent to navigate and interact with complex 3D environments, making strategic decisions in real-time to reach the target object efficiently. The step reward for the agent is defined as:
\begin{equation}
    r(t) = \tanh ( \frac{dis2target(t-1)-dis2target(t)}{max(dis2target(t-1),300)} - \frac{|Ori|}{90 ^{\circ}} )
\end{equation}
where $dis2target(t)$ is the Euclidean distance between the agent and the target at a given timestep $t$ and $|Ori|$ is the absolute orientation error (in degrees) between the agent's current heading and the direction toward the target, normalized by $90^{\circ}$
\subsection{Active Visual Tracking}
\label{app:tracking_task}
Referring to previous works \citep{zhong2024empowering}, we use human characters as an agent player and a continuous action space for agents. The action space contains two variables: the angular velocity and the linear velocity. Angular velocity varies between $-30^{\circ}/s$ and $30^{\circ}/s$, 
while linear velocity ranges from $-1\ m/s$ to $1\ m/s$. In the agent-centric coordinate system, the reward function is defined as:
\begin{equation}
    r = 1- \frac{\vert\rho - \rho^*\vert}{\rho_{max}} - \frac{\vert\theta-\theta^*\vert}{\theta_{max}}
\end{equation}
where $(\rho, \theta)$ denotes the current target position relative to the tracker, $(\rho^*, \theta^*)=(2.5m, 0)$ represents the expected target position, i.e., the target should be $2.5m$ in front of the tracker. The error is normalized by the field of view $(\rho_{max}, \theta_{max})$. During execution, an episode ends with a maximum length of 500 steps, applying the appropriate termination conditions. In the experiment, we adopt the original neural network structure and parameters, as listed in Table ~\ref{offline_tracking_agent} and ~\ref{tab:HyperParam_offline}. 

\begin{figure*}[t]
    \centering
    \includegraphics[width=1\linewidth]{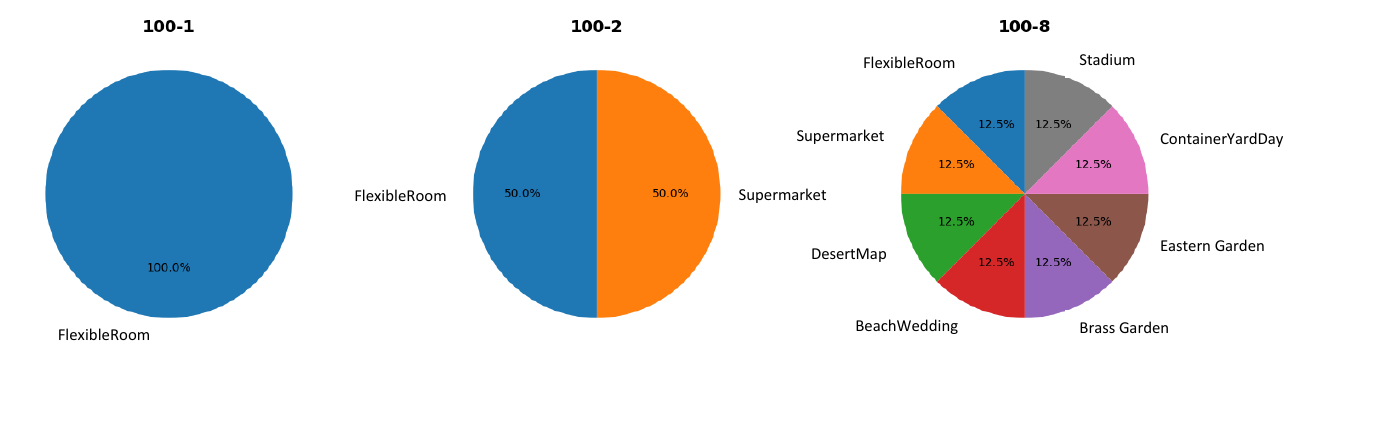}
    \includegraphics[width=1\linewidth]{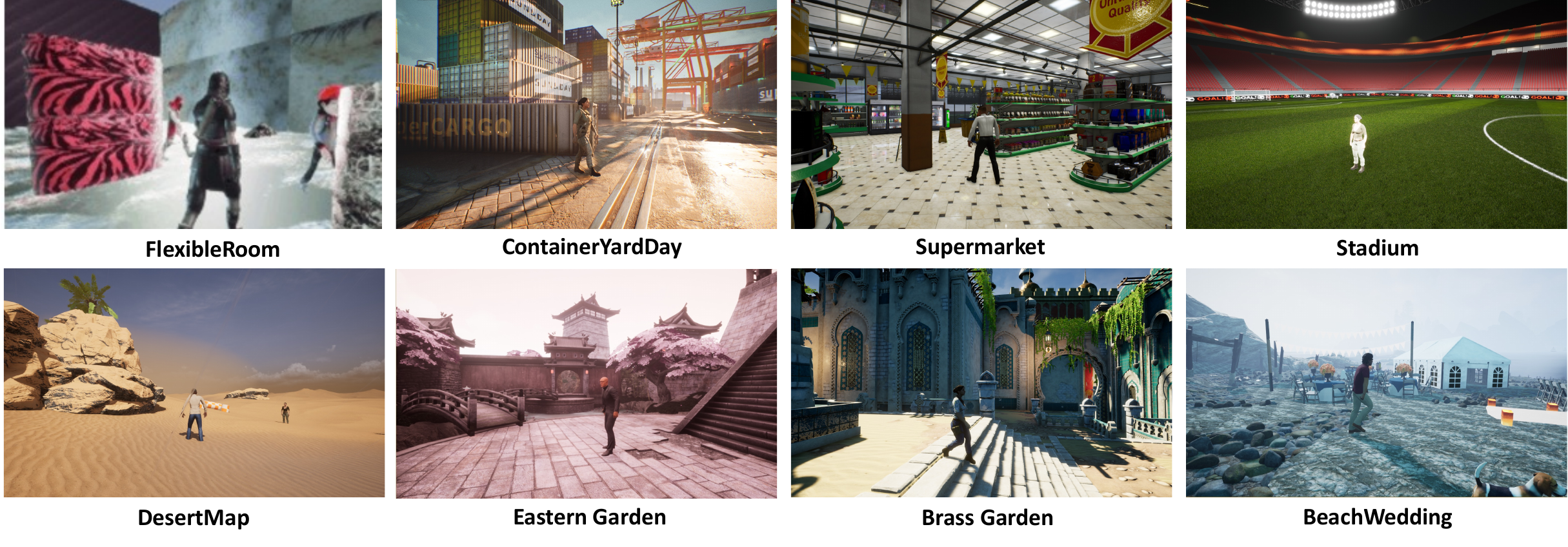}
    \caption{The 8 environments used for collecting the offline dataset.}
    \label{fig:data_distribution}
\end{figure*}

\subsection{Task Configuration in JSON File}
We provide an example of the task configuration JSON file in Figure~\ref{app:json}. Using the JSON file, we can easily set the configuration of the binary, the continuous and discrete action space for each agent, the placement of the binding camera, choose the area to reset, and other hyperparameters about the environments.

\begin{figure*}[t]
  \centering

\begin{tcolorbox}[title= A Json File for Task Configuration ]
\begin{lstlisting}[breaklines=true,texcl=true, escapechar=|]
    "env_name": env_name,
    "env_bin":path-to-binary,
    "env_map": map_name,
    "env_bin_win": path-to-binary(for windows),
    "third_cam": {"cam_id": 0,"pitch": -90,"yaw":0,"roll":0,"height_top_view": 1460.0,"fov": 90},
    "height": 460.0,
    "interval": 1000,
    "agents": {
        "player": {
            "name": ["BP_Character_923"],
            "cam_id": [3],
            "class_name": ["bp_character_C"],
            "internal_nav": true,
            "scale": [1,1,1],
            "relative_location": [20,0,0],
            "relative_rotation": [ 0,0,0],
            "head_action_continuous": {"high": [15,15,15], "low": [-15,-15,-15]},
            "head_action": [ [0,0,0],[0,30,0],[0,-30,0]],
            "animation_action": ["stand","jump","crouch"],
            "move_action": [
            [angular, velocity]
                ...
            ],
            "move_action_continuous": {"high": [30,100],"low": [-30,-100]}
        },
        "animal""{
        ...
        }
        "drone": {
            ...
        }
    },
    ...
    "safe_start": [
        [x,y,z],
        ...
    ],
    "reset_area": [x_min,x_maxin,y_min,y_max,z_min,z_max],
    "random_init": false,
    "env": {"interactive_door": []},
    "obj_num": 466,
    "size": 192555.0,
    "area": 9900.0,
    "bbox": [110.0, 90.0,19.45]
\end{lstlisting}
\end{tcolorbox}
  \caption{An example of the task configuration file in JSON format.}
    \label{app:json}
  \label{figure:task_json}
\end{figure*}


\section{Implementation Details of Agents}
\subsection{Data Collection for Offline RL}
To collect demonstrations for offline reinforcement learning, we use a state-based expert policy and the multi-level perturbation strategy ~\citep{zhong2024empowering} to automatically generate various imperfect demonstrations as the offline dataset. 
For active visual tracking, we employ three distinct datasets for training agents via offline reinforcement learning (Offline RL) algorithms, referred to as \textit{1 Env.}, \textit{2 Envs.}, and \textit{8 Envs}. The detailed composition of each dataset is depicted in Figure~\ref{fig:data_distribution}. For the \textit{1 Env.} dataset, we use only the FlexibleRoom, an abstract environment enriched with diverse augmentation factors, to gather 100k steps of trajectory data. For 2 Envs., we collect 50k step trajectories from FlexibleRoom and an additional 50k steps from the Supermarket environment. The 8 Envs. dataset involves eight different environments, with 12.5k steps collected from each. Therefore, \textbf{the total amount of data in the three datasets is the same (100k) to ensure the fairness of the comparison.}
These dataset configurations aim to highlight the critical role of environment diversity in enhancing the generalization capabilities of embodied AI agents.

\subsection{RL-based Agents}
\label{app:onlineRL}
\begin{figure}[t]
    \centering
    \includegraphics[width=0.99\linewidth]{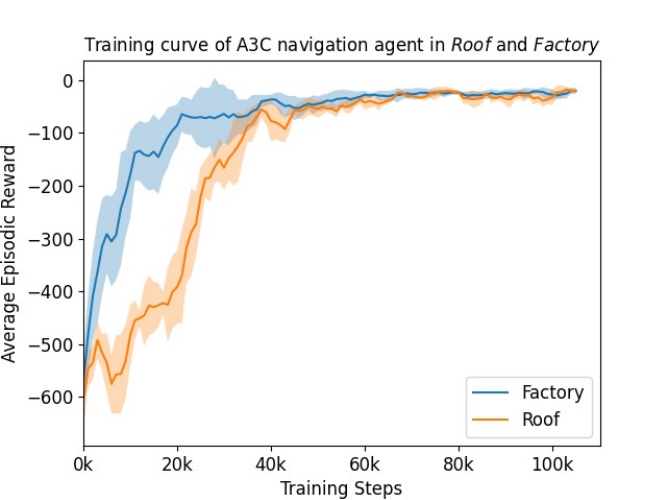}
    \caption{The learning curves for the RL-based navigation agent in two environments: Roof and Factory. We use A3C~\citep{mnih2016asynchronous} to learn the navigation policy via trial-and-error interactions. In the Factory (blue line plot), the number of asynchronous workers is set to 4, while in the Roof environment (orange line plot), the number of asynchronous workers is set to 6. 
    }
    \label{fig:training_curve}
\end{figure}

\textbf{Learning to navigate with online reinforcement learning.} For navigation, we construct an RL-based end-to-end model, using A3C~\citep{mnih2016asynchronous} to accelerate online reinforcement learning in a distributed manner. The model's structure is as follows: a mask encoder extracts spatial visual features from the segmentation mask, which are then passed to a temporal encoder to capture latent temporal information. Finally, the spatiotemporal features, concatenated with the target's relative spatial position, are fed into the actor-critic network to optimize the actor layer for action prediction. The detailed network structure and parameters used in the experiment are listed in Table ~\ref{RL-Based} and ~\ref{tab:HyperParam_nav}. Here, we provide the training curves in \textit{Roof} and \textit{Factory} environments, depicted in Figure~\ref{fig:training_curve}. In the \textit{Factory}, we set the number of workers to 4, while in the \textit{Roof}, the number of workers is set to 6. It can be observed that, for Online RL, the number of workers and the complexity of environments have a significant impact on training efficiency. Looking forward, we anticipate that offline-based algorithms can effectively address the challenges of training efficiency and generalization. 

\begin{table*}[ht]
\centering
\caption{Details the neural network structure of the RL-based agent for the navigation task, where 5$\times$5-32S1 means 32 filters of size 5$\times$5 and stride 1, FC256 indicates the fully connected layer with output dimension 256, and LSTM128 indicates that all the sizes in the LSTM unit are 128.}
\begin{tabular}{c|c|c|c|c|c|c|c|c}
\hline\hline
Module & \multicolumn{8}{|c}{Mask Encoder}\\ \hline
Layer\# & CNN &Pool& CNN &Pool& CNN &Pool& CNN &Pool \\ \hline
Parameters &     5$\times$5-32\emph{S}1 &2-S2& 5$\times$5-32\emph{S}1 &2-S2& 4$\times$4-64\emph{S}1 &2-S2& 3$\times$3-64\emph{S}1 &2-S2  \\ \hline \hline
Module & \multicolumn{2}{|c|}{Temporal Encoder} &\multicolumn{3}{|c|}{ Actor} & \multicolumn{3}{|c}{Critic} \\ \hline
 Layer\# & FC& LSTM & \multicolumn{3}{c|}{FC}& \multicolumn{3}{c}{FC} \\ \hline
 Parameters &256 &128& \multicolumn{3}{c|}{2}&\multicolumn{3}{c}{2} \\ 

 \hline
\end{tabular}
\label{RL-Based}
\end{table*}

\begin{table*}[ht]
    \centering
     \caption{The experiment setting and hyperparameters used for training the RL-based navigation agent.}
   \begin{tabular}{l|c|l|c}
\hline\hline
Name & Value &Name & Value \\ \hline
Learning Rate   & 1e-4 &LSTM update step & 20 \\ 
workers (Roof)   & 6 &LSTM Input Dimension& 256 \\ 
workers (Factory)  & 4 &LSTM Output Dimension& 128\\ 
 Position Input Dimension & 2 &LSTM Hidden Layer size & 1 \\ \hline
 
\end{tabular}
   
    \label{tab:HyperParam_nav}
\end{table*}

\textbf{Learning to track with offline reinforcement learning.} For the tracking task, we adopt an offline reinforcement learning (Offline RL) approach to enhance training efficiency and improve the agent's generalization to unknown environments. Specifically, we build an end-to-end model trained using offline data and the conservative Q-learning (CQL) strategy~\citep{kumar2020conservative}. We adopt the same model structure from the latest visual tracking agent ~\citep{zhong2024empowering}, consisting of a Mask Encoder, a Temporal Encoder, and an Actor-Critic network. Detailed model structures and training parameters are summarized in Table~\ref{offline_tracking_agent} and ~\ref{tab:HyperParam_offline}. Additionally, we provide the model's loss curves under different dataset setups, as shown in Figure ~\ref{fig:offline_curve}.
The model achieves near-convergence within two hours across all dataset setups. To ensure the loss curves stabilize fully, we continued training for an additional three hours, during which no significant further decrease in the loss was observed. A comprehensive evaluation of the model's performance is presented in Tables ~\ref{tab:eval_res} and ~\ref{tab:eval_dis}, highlighting its strong generalization to unseen environments and robustness to dynamic disturbances. The training efficiency, generalization capability, and robustness achieved by offline RL further reinforce our belief that offline RL methods will become a mainstream approach for rapid prototyping and iteration in embodied intelligence systems.

\begin{table*}[ht]
\centering
\caption{Network structure used in the offline RL method \citep{zhong2024empowering}, where 8$\times$8-16S4 means 16 filters of size 8$\times$8 and stride 4, FC256 indicates a fully connected layer with dimension 256, and LSTM64 indicates that all sizes in the LSTM unit are 64. }
\begin{tabular}{c|c|c|c|c|c|c}
\hline\hline
Module & \multicolumn{3}{|c|}{Mask Encoder} & Temporal Encoder & Actor & Critic \\ \hline
Layer\# & CNN & CNN & FC & LSTM & FC & FC\\ \hline
Parameters & 8$\times$8-16\emph{S}4 & 4$\times$4-32\emph{S}2 & 256 & 64 & 2 & 2 \\
\hline
\end{tabular}
\label{offline_tracking_agent}

\end{table*}

\begin{table}[ht]
    \centering
    \caption{The hyperparameters used for offline training and the policy network.}
    \begin{tabular}{l@{\hspace{0.3em}}c@{\hspace{0.8em}}l@{\hspace{0.3em}}c}
    \hline\hline
    Parameter & Value & Parameter & Value \\ \hline
    Learning Rate & 3e-5 & LSTM steps & 20 \\ 
    Discount & 0.99 & LSTM In Dim & 256 \\ 
    Batch Size & 32 & LSTM Out Dim & 64 \\ 
    LSTM Hidden Layers & 1 & & \\ \hline
    \end{tabular}
    \label{tab:HyperParam_offline}
\end{table}

\subsection{VLM-based Agents}
\label{app:prompt}
We built agents with a reasoning framework based on the Large Vision-Language Model. We employ OpenAI GPT-4o as the base model.
System prompt used in the navigation task, as shown in Figure~\ref{app:prompt_navigation}, and system prompt used in the tracking task, as shown in Figure~\ref{app:prompt_tracking}.

\subsection{Human Benchmark for Navigation}
In the navigation task, we incorporated human evaluation as a baseline for comparison to demonstrate the existing gap between the current method and optimal navigation performance. Specifically, \textbf{five male and five female} evaluators participated in the assessment, performing the same navigation tasks under comparable conditions. 

Before each human evaluator began their assessment, we provided a free-roaming perspective to familiarize them with the map structure and clearly conveyed the target's location and image. This ensured that human evaluators had a comprehensive understanding of the environment and the target’s position. During the evaluation, the player was randomly initialized in the environment, and human evaluators used the keyboard to control the agent's movements. Each human evaluator repeated the experiment five times, providing multiple data points to ensure reliability and reduce variability in performance measurements. The termination conditions for the evaluation were identical to those applied to the RL-based agent, ensuring consistency in the comparison.

\begin{figure*}[htbp]
  \centering
\begin{tcolorbox}[title=System Prompt used for active tracking]
\begin{lstlisting}[breaklines=true,texcl=true, escapechar=|]
Objective: 
You are an intelligent tracking agent designed to control the robot to track the person in the view. The first person in your view is your target. You need to provide concrete moving strategie to helo robot tracking the target in the given environment.
Representation details:
1. Moving instructions are concrete actions that the robot can take to adjust its viewpoint and distance to the target. The moving instructions include:
    -move closer: Move the robot closer to the target. This should be chosen when the target is too far away from the robot and there is no obstacle in the way.
    -move further: Move the robot further away from the target. This should be 2chosen when the target is too close to the robot and only part of the target body is visible in the view.
    -keep current: Maintain the current distance and angle between the robot and the target. This is chosen when the target is fully observable in the view and there is enough space in front of both tracker and target without any potential obstacles may cause collision and occlusion.
    -turn left:  Turn the robot to left direction, the target will move towards the right side in next frame. 
    -turn right: Turn the robot to right direction, the target will move towards the left side in next frame. 
Input Understanding:
1.**Image:** We provide a first-person view observation of the robot to help you understand the surrounding environment. The observation is represented as a color image from the tracker's first-person perspective.
Output Understanding:
1. **Moving Strategy:** A temporal reasonable move strategy to adjust the robot viewpoint and distance to achieve robots's long-term tracking task. This should be represented as a concrete moving instructions, the instructions should be choose from "move closer", "move further" ,"keep current", "turn left","turn right". Format - [Keep current].
Strategy Considerations:
1.If the person's horizontal position in the robot's field of view deviates from the center by more than 25\% of the image width, we consider the target to be on one side of the image, otherwise we say the target is near the center. 
Instructions: 
1.Provide ONLY the decision in the [output:] strictly following the format without additional explanations or additional text.
\end{lstlisting}
\end{tcolorbox}
  \caption{System prompt used for tracking.}
    \label{app:prompt_tracking}
\end{figure*}

\begin{figure*}[ht]
  \centering
\begin{tcolorbox}[title=System Prompt used for navigation]
\begin{lstlisting}[breaklines=true]
Objective: 
You are an intelligent navigation agent designed to control the robot to navigate to the target object location based on first-person observation and provide a relative position between the robot and the target. You need to provide an action sequence to help the robot move to the target location.
Representation details:
1. Relative Position: This contains three elements, in the format - [Distance, Direction, Height]. 
    -Distance: The relative distance between the robot and the target object.
    -Direction: The target object's relative direction to the robot, represented in degrees.
1. Actions: These are the movements the robot can perform to adjust its position. The available actions include:
    -Move Forward: Propel the robot forward by 100 centimeter.
    -Move Backward: Propel the robot backward by 100 centimeter.
    -Turn Left: Rotate the robot 15 degrees to the left.
    -Turn Right: Rotate the robot 15 degrees to the right.
    -Jump: Make the robot leap into the air, robot should use this action to jump over obstacles or climb over stairs.
    -Crouch: Lower the robot into a crouching position for 2 seconds, after which it will automatically stand up.
    -Keep Current: Maintain the robot's current position without any movement.
Input Understanding:
1.**Image:** We provide a first-person view observation of the robot to help you understand the surrounding environment. The observation is represented as a color image from the robot's first-person perspective.
2.**Relative Position:** This data provides the target object's relative position to the robot, including the distance, direction, and height. The distance is measured in centimeters, the direction in degrees, and the height in centimeters.
Output Understanding:
1. **Action Sequence:** This is a series of Three continuous actions that the robot should take to navigate toward the target object, in the format - [Action1, Action2, Action3]. Each action should be chosen from the available actions mentioned above.
Strategy Considerations:
1.Assessing Relative Position: Begin by evaluating the target object's relative position in terms of distance, direction, and height to inform the action sequence.
2.Action Combination for Navigation: Utilize the action sequence to create effective combinations, each action will last for 1 seconds. 
3.Obstacle Detection: Leverage the first-person observation to identify obstacles. Based on their location, formulate action sequences that facilitate smooth navigation while avoiding collisions.

Instructions: 
1.Provide ONLY the action sequence in the [output:] strictly following the format -[Action1, Action2, Action3], without additional explanations or additional text.
\end{lstlisting}
\end{tcolorbox}
  \caption{System prompt used for navigation.}
    \label{app:prompt_navigation}
\end{figure*}

\newpage
\vspace{-0.4cm}
\section{Additional Results}
\label{app:res}
\vspace{-0.2cm}
\subsection{Learning Curve}
\label{app:curve}
\begin{figure}[t]
    \centering
    \vspace{-0.2cm}
    \includegraphics[width=1\linewidth]{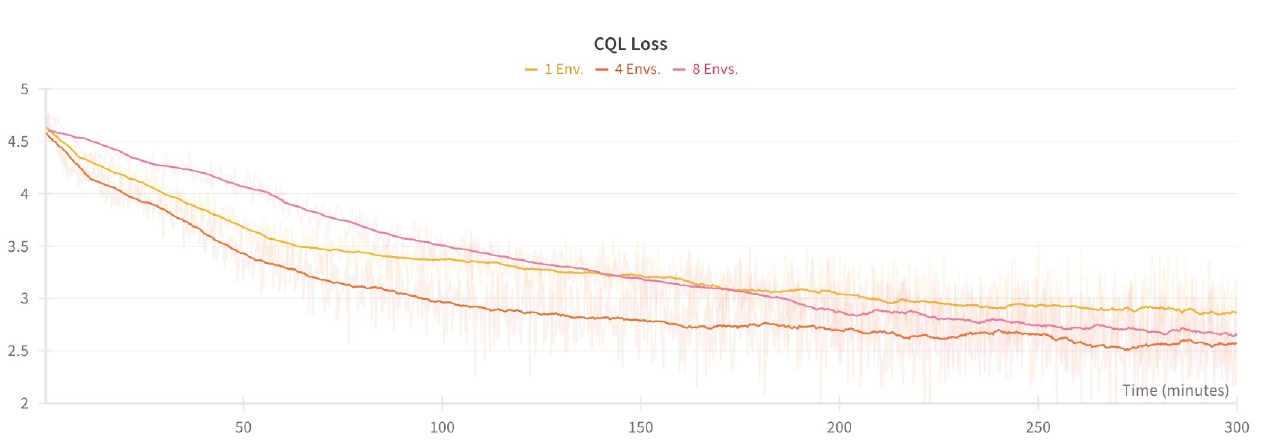}
    \caption{The CQL loss curve during offline training with different offline datasets.}
    \label{fig:offline_curve}
\end{figure}

 We provide the CQL loss curve under the \textit{1 Env., 4 Envs. and 8 Envs.} training setup. As shown in Figure~\ref{fig:offline_curve}, the offline model approaches convergence after two hours, and we continued training for another three hours after nearing convergence, observing no significant further decrease in the loss. Note that the offline training was conducted on an Nvidia RTX 4090 GPU.

\subsection{Testing in 16 Unseen Environments}
\label{app:track16}
We provide the detailed quantitative evaluation results (episodic returns, episode length, success rate) of the RL-based embodied tracking agents across 16 environments, listed in Table~\ref{tab:eval_res}. In each environment, we report the average results over 50 episodes. The results show that in the \textit{Palace Maze}, which contains abundant structural obstacles, the agent's tracking performance was generally weaker compared to the other three categories. In contrast, the agent performed generally better in \textit{Lifelike Urbanity}, characterized by its relatively regular and flat terrain. Additionally, we observed that as the diversity of the training environments increased, the agent's tracking performance improved across all four environment categories. This highlights the positive impact of diverse training data on enhancing the agent's overall tracking effectiveness. We also provide vivid demo videos in \url{https://unrealzoo.notion.site/task-evt}.

\begin{table*}[!t]
    \centering
    \caption{
    Quantitative evaluation results of the offline RL method across 16 environments. The environments are grouped into four categories: Compact Interior, Wildscape Realm, Palace Maze, and Lifelike Urbanity. The table compares the performance of agents trained on different offline dataset settings: 1 Env. (single environment), 2 Envs. (two environments), and 8 Envs. (eight environments). Each cell presents three metrics from left to right: Average Episodic Return (ER), Average Episode Length (EL), and Success Rate (SR). 
   }
   
\begin{tabular}{p{1.4cm}<{\centering}|c|ccc}
\hline
         Category &        Environment Name &  \tabincell{c}{1 Env. \\ ER/EL/SR} &  \tabincell{c}{2 Envs. \\ ER/EL/SR} &  \tabincell{c}{8 Envs. \\  ER/EL/SR} \\\hline
\multirow{4}{*}{\tabincell{c}{Compact \\ Interior}} & Bunker & 241/412/0.56 &  245/391/0.56 &234/429/0.70    \\
                  & StorageHouse & 213 /424 /0.68 &275/449/0.76&170/434/0.64 \\
                  &  SoulCave & 229/402/0.60 &252/422/0.56&206/405/0.58 \\
                  & UndergroundParking & 179/391/0.56 &250/424/0.62&184/410/0.60  \\ \hline
\multirow{4}{*}{\tabincell{c}{Wildscape \\ Realm}} & Desert Ruins & 209/392/0.54 
                  &293/449/0.70&277/453/0.70\\
                  & GreekIsland & 245/411/0.62 &264/423/0.64
                 &257/466/0.78\\
                  & SnowMap & 204/399/0.62 &322/456/0.78 &278/474/0.86\\
                  & RealLandscape & 171 /383/0.42 &225/372/0.44&223/444/0.70 \\ \hline
\multirow{4}{*}{\tabincell{c}{Palace \\ Maze}} &  WesternGarden & 230/403/0.54 &  
                209/408/0.54&296/472/0.82 \\
                  & TerrainDemo & 232/411/0.56 & 233/403/0.56&192/411/0.56\\
                  & ModularGothicNight & 190/360/0.52 & 244/423/0.62&272/456/0.76\\
                  &ModularSciFiSeason1 & 168/365/0.42 &172/354/0.42&211/393/0.48\\ \hline
\multirow{4}{*}{\tabincell{c}{Lifelike \\ Urbanity}} &SuburbNeighborhoodDay & 224/422/0.64 &328/457/0.72&242/457/0.76\\
                  &DowntownWest & 296/460/0.78 &317/456/0.76		&292/469/0.86 \\
                  & Factory & 278/434/0.64 &291/452/0.74&249/435/0.64\\
                  & Venice & 295/441/0.70 &323/448/0.82&294/474/0.84\\ \hline
\end{tabular}
    \label{tab:eval_res}
\end{table*}

\begin{table*}[tb]
    \centering
    \caption{
   Quantitative evaluation results of the tracking agents across 4 different category environments with \textbf{4 distractors (4D), 8 distractors (8D), and 10 distractors (10D)} respectively. The table compares the performance of agents trained on different offline dataset settings: 1 Env. (single environment), 2 Envs. (two environments), and 8 Envs. (eight environments). Each cell presents three metrics from left to right:  Average Episodic Return (ER), Average Episode Length (EL), and Success Rate (SR).
   }
\begin{tabular}{p{1.4cm}<{\centering}|c|ccc}
\hline
         Category &        Environment Name &  \tabincell{c}{1 Env. \\ ER/EL/SR} &  \tabincell{c}{2 Envs. \\ ER/EL/SR} &  \tabincell{c}{8 Envs. \\  ER/EL/SR} \\ \hline
\multirow{3}{*}{\tabincell{c}{Compact \\ Interior}} &           StorageHouse (4D)& 117/343/0.40 & 181/375/0.52   &190/428/0.62\\
&           StorageHouse (8D)& 143/341/0.34 & 151/338/0.44   &165/366/0.49\\
                  & StorageHouse (10D)& 81/324/0.36 &109/331/0.42& 107/357/0.50\\ \hline
                 
\multirow{3}{*}{\tabincell{c}{Wildscape \\ Realm}} 
                    & DesertRuins (4D)& 317/469/0.72
                  &333/456/0.70&354/466/0.74\\ 
                  & DesertRuins (8D)& 213/406/0.50
                  &316/445/0.58&267/444/0.68\\
                  & DesertRuins (10D)& 188/390/0.44 
                  &252/382/0.50&253/447/0.64\\
                  \hline
\multirow{3}{*}{\tabincell{c}{Palace \\ Maze}} &  TerrainDemo (4D)&  221/398/0.44 & 286/454/0.65 & 312/460/0.77 \\
&  TerrainDemo (8D)& 211/384/0.39 & 239/412/0.49 & 252/420/0.52 \\
                  & TerrainDemo (10D)& 189/377/0.36 & 232/404/0.48&224/429/0.66\\
                  \hline

\multirow{3}{*}{\tabincell{c}{Lifelike \\ Urbanity}} 
&SuburbNeighborhoodDay (4D)     
  &192/407/0.46&256/381/0.50&265/392/0.60\\
  &SuburbNeighborhoodDay (8D)     
  &131/325/0.36&229/369/0.48&247/385/0.56\\
&SuburbNeighborhoodDay (10D)     
  &162/355/0.44&180/340/0.40&165/376/0.44\\
                 \hline

\end{tabular}
    \label{tab:eval_dis}
\end{table*}

\subsection{Testing in Social Tracking Scenarios}
\label{app:social}

We select 4 environments from different categories as the testing environments, including StorageHouse, DesertRuins, TerrainDemo, and SuburbNeighborhoodDay. We test the distraction robustness of the social tracking agents by adding different numbers of distractors (4, 8, 10) in the environment. The distractors randomly walk around the environment, which may produce various unexpected perturbations to the tracker, such as visual distractions, occlusion, or blocking the tracker's path. As shown in Table~\ref{tab:eval_dis}, the tracking performance of the three agents steadily decays with the increasing number of distractors.

\subsection{Cross-platform Evaluation}
To evaluate the generalization ability of policies trained in UnrealZoo, we conducted cross-platform experiments in both simulation and real-world settings. During our investigation, we found that many existing simulation environments ~\cite{puig2023habitat,thor2017, puig2018virtualhome} lack flexibility in task customization, making it difficult to add or modify task definitions. Among the available options, we selected ThreeDWorld~\cite{gan2020threedworld} for its minimal migration overhead.

\begin{figure}
    \centering
    \includegraphics[width=\linewidth]{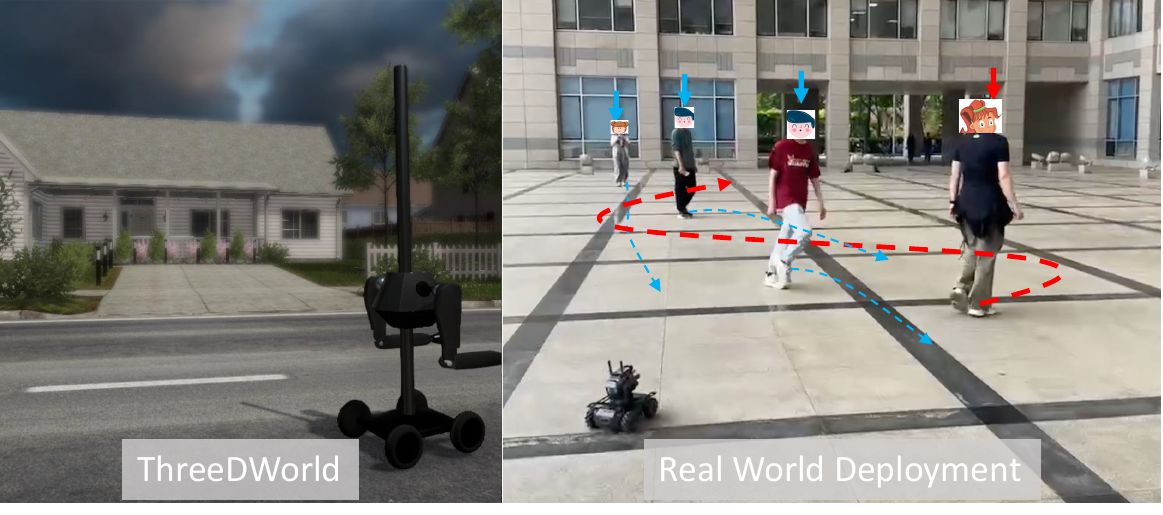}
    \caption{From left to right: 1)We utilize ThreedWorld as the comparison simulator and use its built-in agent ``Magnebot'' as the target, evaluating the active tracking performance. 2) We deploy our trained policy in a wheel-based robot with distractors, evaluating the active tracking performance.}
    \label{fig:cross_platform}
\end{figure}

For evaluation, we used the publicly available "Suburb Scene 2023" map and designated the robot Magnebot as the tracking target, as shown in Figure~\ref{fig:cross_platform}. We tested two policies trained under different environment settings: a single environment (1-Env) and eight diverse environments (8-Envs). Due to limitations in ThreeDWorld, specifically, the inability to retrieve the absolute position of Magnebot, we could not compute the Average Reward (AR) metric. Instead, we report the episode length (EL) and success rate (SR) as alternative indicators of tracking performance.

Beyond simulation, we further validated the trained policies in the real world, illustrated in Figure~\ref{fig:cross_platform}. In this scenario, the target person walks along an S-shaped path in an open area with various distractors. We conducted five trials for each policy. Since all trials were successful, the success rate alone could not reflect performance differences. Additionally, absolute position data was unavailable in the real-world setting, so we introduced two alternative evaluation metrics:
\begin{itemize}
\item \textbf{Average Deviation (AD)}: the average pixel offset between the target's bounding box center and the image center.
\item \textbf{IoU}: the Intersection over Union between the current target bounding box and the initial bounding box.
\end{itemize}

These metrics assess the agent’s ability to maintain consistent and centered tracking under dynamic motion and environmental interference. The final results are listed in Table~\ref{tab:cross_platform}. Overall, the experimental findings highlight that training with diverse environments in UnrealZoo significantly improves the policy's generalization ability and robustness.

\begin{table}
\small
    \centering
    \vspace{-0.2cm}
    \caption{Cross-platform evaluation on active visual tracking task}
    \vspace{-0.3cm}
\begin{tabular}{c|cc}
\hline
      Platform  & ThreeDWorld & Real World \\ 
                & EL/SR       & AD/IoU     \\ \hline
1 Env.          & 377/0.65    & 764/328    \\
8 Envs.         & 493/0.8     & 833/393    \\
\hline
\end{tabular}
\vspace{-0.2cm}
\label{tab:cross_platform}
\vspace{-0.5cm}
\end{table}

\section{Limitations and Discussions.} While our proposed environment provides diverse and complex scenarios for visual navigation, tracking, and other embodied vision tasks, it currently has some limitations that should be overcome in future work:
1) Limited Physical Simulation Fidelity. The current version focuses on visual rendering and interaction rather than the precision of physical interactions for control. We will consider enhancing physics engine integration for more accurate and realistic physical interactions.
2) Licensing Restrictions and Shareability. Due to marketplace content licensing restrictions, we are unable to open-source the complete environment source code for more in-depth customization. To address this, we will develop alternative scenes based on free assets or negotiate special licensing agreements with asset creators, and extend the UnrealCV command system to support more advanced customization in the released binary.
3) Limited Interactions. Interaction capabilities are primarily limited to pre-defined objects, and multi-agent interactions remain relatively basic, lacking complex social dynamics. We will extend to more complex interaction tasks such as object manipulation, tool use, and environmental modification, and implement more sophisticated social agent behavior models, including dialogue and emotional expression.